\documentclass[MS]{macro/neu_msthesis}

\usepackage{physics}
\usepackage{siunitx}
\usepackage{graphics} 
\usepackage{booktabs}
\usepackage{enumitem}
\usepackage{graphicx} 
\usepackage{epsfig} 
\usepackage{lipsum} 
\usepackage{amsmath, bm} 
\usepackage{amssymb}  
\usepackage{upgreek}
\usepackage{url}
\usepackage{breakurl}
\usepackage{caption}
\usepackage{subcaption}
\usepackage{pgfplots}
\usepackage{floatrow}
\usepackage{float}
\usepackage[ruled,linesnumbered]{algorithm2e}
\usepackage{multicol}
\usepackage{tikz}
\usetikzlibrary{shapes, arrows}
\usepackage{amsmath}
\newfloatcommand{capbtabbox}{table}[][\FBwidth]

\title{NMPC-based Unified Posture Manipulation and Thrust Vectoring for Agile and Fault-Tolerant Flight of a Morphing Aerial Robot}

\author{Shashwat Pandya}

\dept{Mechanical and Industrial Engineering}

\degree{Master of Science}

\degreename{Robotics}

\field{Robotics}

\submitdate{April 2025}

\numberofmembers{3}

\principaladviser{Dr. Alireza Ramezani}
\firstreader{Dr.Rifat Sipahi}

\chairman{Dr. Yingzi Lin}

\dean{Sagar Kamarthi}

\usepackage{amsfonts}			
\usepackage{times}		

\newcommand{\ifno}[1]{}

\usepackage{multirow}
\makeindex
\usepackage{makeidx}
\usepackage{acronym}

\usepackage{url}
\usepackage{graphicx}

\ifnum\pdfoutput>0
\usepackage[pdftex,
bookmarks=true,
bookmarksnumbered=true,
hypertexnames=false,
breaklinks=true           
]{hyperref}
\else
\usepackage[hypertex,
bookmarks=true,
bookmarksnumbered=true,
hypertexnames=false,
breaklinks=true           
]{hyperref}
\fi

\hypersetup{				
pdfauthor = {\authorRef},
pdftitle = {\titleRef},
pdfsubject = {\expandafter{\degreeRef} thesis submitted to Northeastern University},
pdfkeywords = {Multimodal robots,M4,Traversability,3D path planning}
pdfcreator = {LaTeX with hyperref package},
pdfproducer = {dvips + ps2pdf}}

\begin{document}

\pdfbookmark[1]{Cover}{cover}

\titlepage

\begin{frontmatter}

\begin{dedication}
To my Lord and my family.
\end{dedication}

\pdfbookmark[1]{Table of Contents}{contents}
\tableofcontents
\listoffigures
\newpage\ssp
\listoftables


\chapter*{List of Acronyms}
\addcontentsline{toc}{chapter}{List of Acronyms}

\begin{acronym}
\acro{M4}{ Multi-Modal Mobility Morphobot}.
\label{acro:M4}

\acro{NMPC}{Non-linear Model Predictive Control}.
\label{acro:NMPC}

\acro{UAVs}{Unmanned Aerial Vehicles}.
\label{acro:UAVs}

\acro{QP}{Quadratic Programming}.
\label{acro:QP}

\acro{COBRA}{Crater Observing Bio-inspired Rolling Articulator}.
\label{acro:COBRA}

\acro{RL}{Reinforcement Learning}.
\label{acro:RL}

\acro{PID}{Proportional Integral Derivative}.
\label{acro:PID}

\acro{FTC}{Fault-Tolerant Control}.
\label{acro:FTC}

\acro{EKF}{Extended Kalman Filter}.
\label{acro:EKF}

\acro{IMU}{Inertial Measurement Unit}.
\label{acro:IMU}

\acro{SNNs}{Spiking Neural Networks}.
\label{acro:SNNs}

\acro{MPC}{Model Predictive Control}.
\label{acro:MPC}

\acro{CoG}{Center of Gravity}.
\label{acro:CoG}

\acro{DOF}{Degree oF Freedom}.
\label{acro:DOF}

\acro{SINDy}{Sparse Identification of Nonlinear Dynamics}.
\label{acro:SINDy}

\acro{SMC}{Sliding Mode Control}.
\label{acro:SMC}

\acro{LQR}{Linear Quadratic Regulator}.
\label{acro:LQR}

\acro{LoE}{Loss of Effectiveness}.
\label{acro:LoE}

\end{acronym}


\begin{acknowledgements}

I extend my heartfelt gratitude to all the members of the SiliconSynapse Lab at Northeastern University for their unwavering support throughout my thesis journey. I am especially grateful to Prof. Alireza Ramezani, my primary advisor for their exceptional mentorship, belief in my abilities, and invaluable guidance. Their expertise and dedication have inspired me to strive for excellence in my research. I would also like to acknowledge the contributions of Adarsh Salagame, Shreyansh Pitroda, Ioannis Mandralis (CalTech), Kaushik Venkatesh Krishnamurthy, Bibek Gupta and Chenghao Wang, whose support and collaboration have enriched my work. Special thanks go to Prof. Rifat Sipahi for serving as my mechanical engineering department co-adviser and thesis reader. I am deeply thankful for the support of my parents, whose unwavering love and encouragement have been the driving force behind my achievements. I am excited about the future possibilities it opens up in the field of robotics and autonomous systems.

\end{acknowledgements}


\begin{abstract}

This thesis presents a unified control framework for agile and fault-tolerant flight of the Multi-Modal Mobility Morphobot (M4) in aerial mode. The M4 robot is capable of transitioning between ground and aerial locomotion. The articulated legs enable more dynamic maneuvers than a standard quadrotor platform. A nonlinear model predictive control (NMPC) approach is developed to simultaneously plan posture manipulation and thrust vectoring actions, allowing the robot to execute sharp turns and dynamic flight trajectories. The framework integrates an agile and fault-tolerant control logic that enables precise tracking under aggressive maneuvers while compensating for actuator failures, ensuring continued operation without significant performance degradation. Simulation results validate the effectiveness of the proposed method, demonstrating accurate trajectory tracking and robust recovery from faults, contributing to resilient autonomous flight in complex environments.


\end{abstract}

\end{frontmatter}


\pagestyle{headings}

\chapter{Introduction}

UAvs have revolutionized applications ranging from surveillance and disaster response to planetary exploration. Their ability to rapidly deploy, navigate complex environments, and access otherwise unreachable terrain makes aerial robots uniquely valuable across civilian and scientific domains. However, many aerial platforms remain constrained by limited agility, poor resilience to actuator failures, and over-reliance on simplified dynamics and control architectures.

Traditionally, quadrotor UAVs have dominated the field due to their mechanical simplicity and intuitive control. Yet, their fixed-thrust configurations and symmetric geometry make them vulnerable to actuator degradation or rotor failures, significantly reducing their performance. Additionally, conventional aerial robots offer limited maneuverability in constrained environments due to insufficient thrust vectoring or redundant actuation.

The exploration of planetary environments, such as Mars, exemplifies the limitations faced by existing robotic systems due to unpredictable terrain, rugged landscapes, and limited communication bandwidth. Traditional wheeled rovers encounter obstacles like large rocks, craters, or steep slopes, while conventional aerial robots struggle with agility and robustness in response to actuator faults. Consequently, platforms capable of dynamic posture adaptation during mid-flight provide critical advantages by offering robust solutions to unforeseen terrain challenges, thus enabling more efficient planetary exploration.

To overcome these limitations, recent research has increasingly focused on developing multimodal robotic systems capable of seamlessly transitioning between various forms of locomotion, such as wheeled, legged, aerial, and hybrid modalities. Multimodal robots leverage the strengths of different mobility modes to enhance locomotion plasticity, resilience, and adaptability across diverse terrains and complex environments.


For instance, the M4 robot employs appendage re-purposing to dynamically adapt its locomotion strategy, enabling efficient navigation across challenging terrains by utilizing wheels, thrusters, and legs interchangeably for flying, rolling, crawling, and balancing \cite{M4_nature}. Recent advancements on the M4 platform include autonomous 3D path planning and the integration of efficient path tracking algorithms tailored for complex indoor environments, underscoring its suitability for versatile deployment in exploration and search-and-rescue scenarios \cite{sihite2023demonstrating, sihite2022efficient}. Additionally, self-supervised learning approaches have been implemented to autonomously estimate the cost of transport using visual inputs, significantly enhancing the robot's ability to select energetically optimal paths across varying terrains such as grass and paved surfaces, while maintaining low computational overhead suitable for onboard deployment on compact robotic computing units \cite{gherold_self-supervised_2024}. Further studies have expanded upon these capabilities, focusing on efficient multi-modal navigation specifically tailored for extraterrestrial exploration contexts, such as Mars surface traversals, and initial implementations of autonomous navigation methods onboard the M4 platform \cite{ben, filip}. Contributions such as efficient morphing micro aerial vehicle design and thruster-assisted surface mobility highlight M4’s relevance to Mars exploration efforts \cite{ramezani2022efficient, ramezani2022thruster}. In parallel, intellectual property efforts have solidified the M4’s role as a next-generation morpho-functional robotic concept through recent U.S. patent applications \cite{ramezani2023multi, ramezani2023morpho}. Most recently, the ATMO (Aerially Transforming Morphobot) concept has been introduced to enable dynamic transitions between ground and aerial modes, showcasing transformative capabilities that further extend M4’s operational envelope \cite{mandralis2025atmoaeriallytransformingmorphobot}.

Similarly, Husky Carbon is a sophisticated legged-aerial platform capable of combining wheeled and thruster-assisted legged mobility to traverse diverse landscapes effectively. It integrates aerial capabilities, enabling it to overcome large obstacles and steep slopes through dynamic posture manipulation and thrust vectoring, effectively mimicking wing-assisted inclined running observed in certain birds \cite{ramezani_generative_2021, salagame2022letter, salagame2024quadrupedal, krishnamurthy_thruster-assisted_2024}. Recent studies on Husky Carbon include the development of optimization-free control frameworks utilizing momentum observers for efficient ground reaction force estimation, significantly simplifying onboard computational requirements while maintaining robust locomotion on complex terrains \cite{krishnamurthy_optimization_2024}. Additional research demonstrated steep slope walking using a reduced-order modeling approach combined with quadratic programming-based model predictive control, illustrating the robot's advanced agility on highly inclined surfaces \cite{krishnamurthy_enabling_2024}. Moreover, further investigations explored narrow-path dynamic walking, integrating posture manipulation and thrust vectoring to enhance maneuverability and stability in constrained environments \cite{krishnamurthy2024narrow}. These efforts collectively highlight Husky Carbon's potential in performing robust locomotion tasks across unpredictable terrains, facilitated by advanced path-planning techniques such as integrated probabilistic road maps (PRM) and reference governors (RG), along with optimization-free ground contact force constraint satisfaction \cite{sihite2021optimization, PRM-MM}.


Extending the principles of adaptive multimodal locomotion to bipedal systems, the thruster-assisted bipedal robot Harpy showcases remarkable agility and robustness by leveraging dynamic posture manipulation in combination with thrusters. This integration allows Harpy to perform advanced maneuvers previously unattainable by traditional bipedal platforms, such as dynamic wall-jumping, steep slope walking, and capture point control to stabilize locomotion on inclined surfaces, thereby demonstrating superior resilience and adaptability in complex and dynamic environments \cite{dangol2021control, sihite2024posture, pitroda_capture_2024}. Advanced control strategies, including hybrid zero dynamics (HZD), QP, and conjugate momentum-based thruster force estimation, have been developed to further enhance the robustness and efficiency of thruster-assisted locomotion \cite{dangol2021hzd, pitroda_conjugate_2024, pitroda_enhanced_2024}. Specifically, QP-based optimization incorporating contact constraints has been utilized to develop thruster-assisted slope walking controllers inspired by avian locomotion mechanisms \cite{pitroda_quadratic_2024}. Additional theoretical work has also explored the fundamental limits of thruster-aided bipedal motion, including legged locomotion on inclined terrains and feasibility analysis for dynamic terrain negotiation \cite{dangol2020towards, changaoss1}.

\begin{sloppypar}
Additional hybrid multimodal systems include flying-driving robots like FSTAR and HyFDR, capable of combining efficient terrestrial locomotion with aerial mobility, particularly suited for exploration, package delivery, and search-and-rescue missions in cluttered environments \cite{gefen_flying_2022, sharif_energy_2018}. Robots such as ANYmal with wheeled-leg integration further enhance locomotion efficiency and maneuverability, demonstrating applications in planetary exploration and emergency response \cite{ANYmal_Wheeled, anymal_lune}. Furthermore, robots such as DRAGON and SPIDAR emphasize aerial manipulation and air-ground amphibious capabilities, enabling complex object manipulation and seamless transition between terrestrial and aerial navigation \cite{zhao_versatile_2023, zhao_design_2023}.
\end{sloppypar}
Collectively, these multimodal robotic systems have demonstrated significant promise across a range of challenging tasks, including planetary exploration, search-and-rescue operations, dynamic obstacle navigation, and high-maneuverability aerial reconnaissance, thus setting a robust foundation for future robotic developments in complex and unpredictable environments.

Another notable development in terrain-adaptive mobility is the morphofunctional robot COBRA robot, a snake-like platform capable of transitioning between traditional gaits and passive tumbling for rugged terrain traversal. Originally proposed to address challenges in lunar crater exploration, COBRA integrates articulated modules that enable both lateral undulation and controlled rolling, thereby allowing it to navigate steep slopes and uneven surfaces with enhanced stability. Experimental and simulation-based studies have demonstrated closed-loop control of its tumbling dynamics via posture manipulation, enabling directional control during descent \cite{salagame_dynamic_2024, salagame_validation_2024}. Advanced modeling and planning frameworks have also been explored to achieve contact-implicit loco-manipulation through lateral rolling, extending COBRA’s functional envelope to include object interaction via body motion \cite{salagame_non-impulsive_2024}. Moreover, hierarchical RL controllers have enabled large-scale adaptive navigation by dynamically switching among gait strategies in unknown environments \cite{jiang_hierarchical_2023}. These strategies are further enhanced by tactile perception systems that allow COBRA to adapt its motion based on environmental contact feedback, significantly improving terrain responsiveness in cluttered settings \cite{jiang_snake_2024}. To address the sim-to-real gap, recent work has applied RL-based model matching to refine the physical simulation parameters of COBRA, improving real-world locomotion accuracy \cite{salagame_reinforcement_2024}. Recent work has also introduced a posture-based heading control framework for COBRA, enabling it to perform real-time obstacle avoidance during passive tumbling by manipulating internal joint configurations to influence roll curvature and forward direction \cite{salagame_heading_2024}. This expands COBRA’s control authority despite its energy-efficient passive rolling behavior and has been validated through hardware testing. Collectively, these advancements illustrate COBRA’s viability as a robust platform for confined-space exploration, disaster-response, and planetary surface mobility in extreme terrains.

Recent advances in aerial robotics also attempt to address the shortcomings of standard robotic platforms through various innovative architectures, including tilt-rotor designs \cite{ruihang2020}, fully actuated tri-rotors \cite{kara2012}, and flying humanoids such as iRonCub \cite{gabriele2023}. These systems enhance the robustness and maneuverability of aerial platforms by introducing mechanical degrees of freedom and over-actuated configurations. For instance, tilting quadcopters can vector thrust to improve maneuverability \cite{nemati2016stability}, while humanoid aerial systems utilize articulated limbs for in-flight stabilization and dynamic reconfiguration \cite{gabriele2023}.

In addition to over-actuated rotorcraft and humanoid flyers, recent work has explored morphing-wing platforms like Aerobat, which achieve agility through dynamic structural reconfiguration rather than conventional thrust-vectoring. Designed with bio-inspired articulating wings, Aerobat performs advanced maneuvers such as banking turns and bounding flight, leveraging high degrees of freedom in its morphing geometry. A recent study demonstrated how wing-induced shift in structural response could be exploited to execute coordinated banking maneuvers using collocation-based optimization \cite{gupta2024aim}. In parallel, the system has been experimentally validated for intermittent bounding flight, mimicking avian gait cycles through synchronized wing collapse and actuation-assisted stabilization \cite{gupta_bounding_2024}. To support hovering and orientation control, a hybrid platform combining flapping wings and multi-rotors has also been developed, using state observers to maintain closed-loop control even in the presence of actuation uncertainty \cite{dhole_hovering_2023}. Complementing these efforts, a fluid–structure interaction (FSI) modeling framework has been proposed to fine-tune Aerobat’s wing response to aerodynamic forces, enabling predictive tracking of complex 3D trajectories in simulation \cite{gupta_modeling_2024}. Collectively, these contributions position Aerobat as a dynamic morphing aerial system capable of executing biologically inspired flight behaviors with fine-grained control authority.

Despite such advancements, most UAV control solutions still heavily rely on simplified frameworks, such as PID controllers, which, despite their computational simplicity, often fail in scenarios involving nonlinear dynamics, actuator faults, or significant external disturbances. Advanced scenarios like high-speed maneuvering through cluttered environments or sustained flight post-actuator failures require more sophisticated control approaches. Limitations of existing methods become pronounced when precise control must be maintained despite severe hardware degradation or environmental unpredictability.

Articulated limbs offer a promising solution by introducing additional degrees of freedom that significantly influence aerial robot stability and control. However, leveraging these capabilities within real-time flight control frameworks remains underexplored, presenting a substantial opportunity for innovative research in aerial robotics.

\begin{figure}[!htbp] \centering \includegraphics[width=0.8\linewidth]{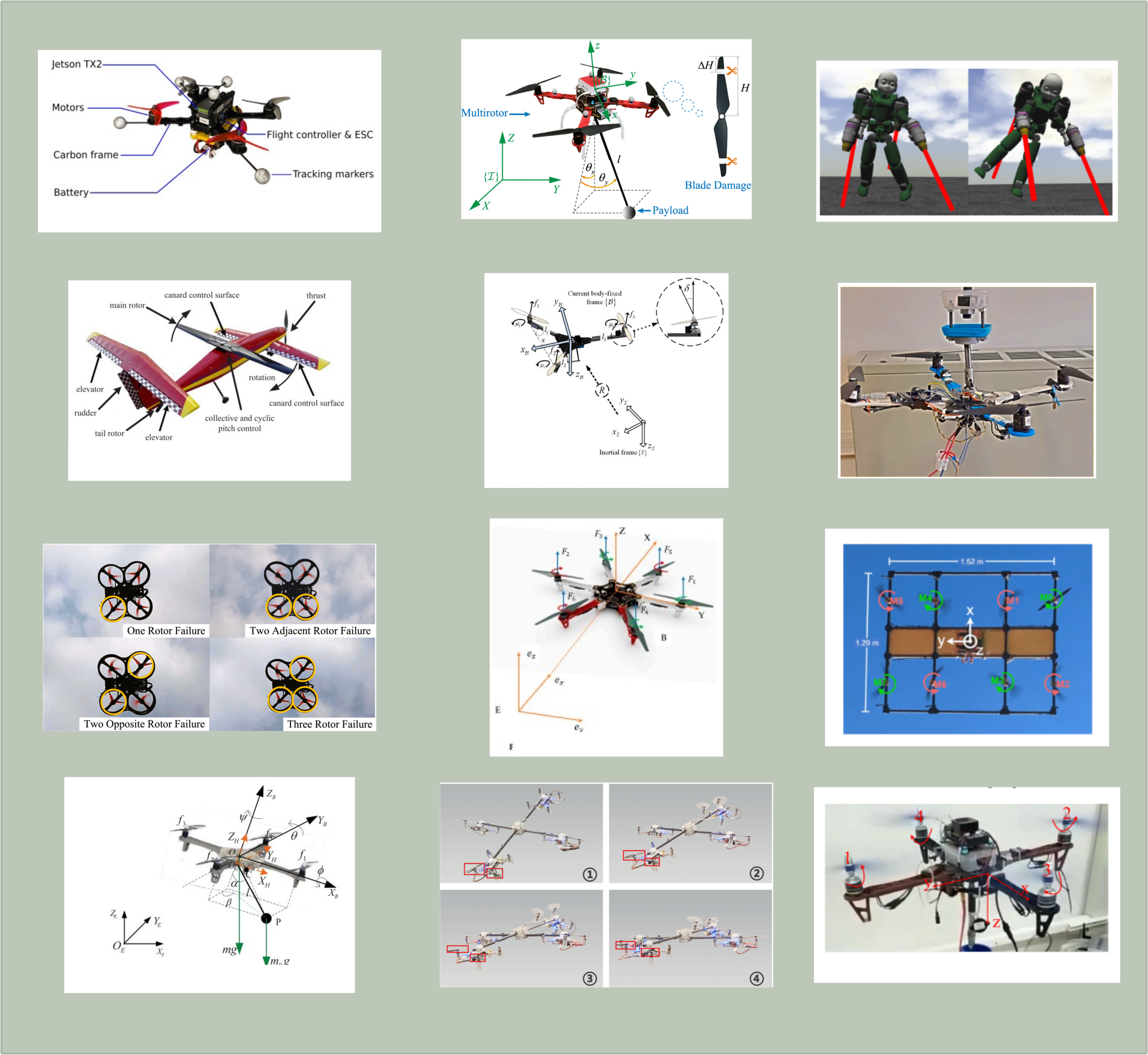} \caption{Various examples of aerial robotic systems.\cite{nan2022nonlinear,gabriele2023, wang2023, hao2022fault, yao2024, ke2023uniform, shen2021,mazare2024robust, nguyen2019, oconnell2024learning,yu2024fault,ahmadi2023}} \label{fig:morph} \end{figure}

\begin{figure}[!htbp] \centering \includegraphics[width=\textwidth]{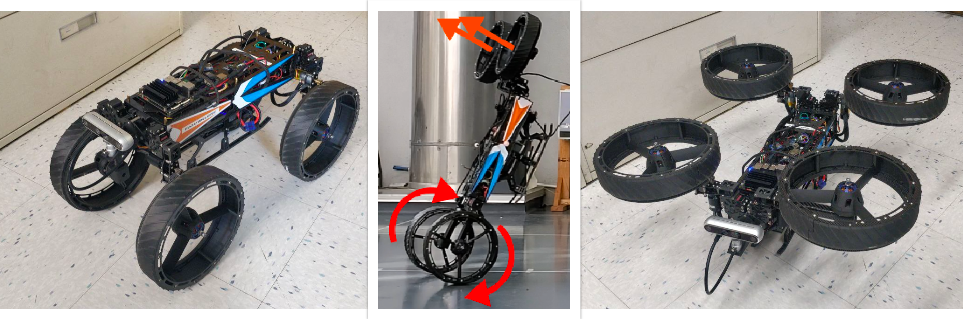} \caption{The M4 robot in rover, segway, and aerial configurations \cite{filip}.} \label{fig:morph} \end{figure}

This thesis specifically explores the aerial capabilities of the M4, a morphing robot equipped with articulated limbs capable of dynamic mid-air reconfiguration. Unlike standard quadrotors, M4 exploits posture manipulation to dynamically modify its center of mass, aerodynamic profile, and thrust vector orientation, significantly expanding its maneuverability envelope. Initially designed for multi-modal exploration, the aerial mode of M4 presents novel control challenges related to exploiting joint articulation for achieving robust, fault-tolerant, and agile maneuvers.

Articulated limbs add mechanical complexity but simultaneously provide unique advantages such as aerodynamic adaptability and real-time mass redistribution. Inspired by biological organisms like birds, which dynamically position limbs to stabilize and maneuver during flight, aerial robotic systems with articulated limbs can leverage similar mechanisms to adjust their aerodynamic profiles and thrust vectors. However, such control requires addressing substantial challenges, including increased computational demands, complex state estimation, and real-time optimization of posture and thrust vectors.

Recent research by Mandralis et al. \cite{mandralis2023minimum} has investigated M4's morphing capabilities through combined posture control and thrust vectoring for minimum-time trajectories under spatial constraints. The study demonstrated through numerical simulations that posture manipulation significantly improves the robot’s ability to navigate narrow geometries and enhances maneuverability. They formulated optimal control problems solved using trapezoidal collocation methods, validating their strategy in two scenarios: maneuvering through a spatially varying tunnel and tracking predefined trajectories under tight spatial constraints. The results underscored the effectiveness of combined posture and thrust vectoring for reducing maneuver times and facilitating challenging trajectory execution. Nevertheless, this integrated control paradigm remains largely unexplored. Building upon this foundation, this thesis aims to further develop posture and thrust manipulation strategies to enhance aerial multi-modal robot agility and adaptability.

Specifically, we propose a unified NMPC framework that integrates posture manipulation with thrust vectoring. NMPC is particularly advantageous due to its explicit handling of nonlinear system dynamics, actuator constraints, and dynamic objectives. Unlike conventional control strategies, NMPC implicitly manages actuator faults by dynamically reallocating thrust and exploiting limb articulation, removing the need for predefined recovery strategies. Its inherent adaptability thus makes NMPC particularly suitable for reliable and agile flight under unpredictable conditions.

Compared to existing fault-tolerant strategies, such as adaptive sliding mode control \cite{mallavalli2019fault}, hybrid learning-based compensation \cite{oconnell2024learning}, or tilt-rotor allocation methods \cite{ryll2012}, our approach emphasizes generality, and implicit fault recovery. Rather than explicitly modeling each fault scenario separately, redundancy is embedded directly into the NMPC formulation, utilizing joint articulation as an auxiliary control input.

The specific goals of this thesis include: \begin{itemize} \item Developing a reduced-order model of the aerial M4 configuration that captures interactions between joint articulation and body dynamics; \item Formulating an NMPC framework integrating posture manipulation and thrust vectoring for fault-tolerant and agile trajectory tracking; \item Implementing and validating the NMPC control strategy using high-fidelity Simscape simulations under varied flight conditions; \item Demonstrating successful fault recovery and agile maneuvering in simulations employing combined posture manipulation and thrust vectoring. \end{itemize}

The broader motivation behind this work is to bridge the gap between mechanical redundancy and algorithmic fault-tolerance by introducing control strategies that exploit all available degrees of freedom, including limb articulation. This research contributes significantly toward the ongoing evolution of aerial robots, transforming them from passive, limited-capability platforms into resilient, reconfigurable systems capable of complex missions in cluttered or unpredictable environments.

Ultimately, the developed framework moves us closer to deploying robust aerial robotic systems capable of effectively addressing mission-critical scenarios like planetary exploration, dense forest search-and-rescue operations, and high-speed navigation in GPS-denied indoor environments, where reliability, agility, and reconfigurability are paramount.

This thesis is organized as follows: Chapter 2 provides a comprehensive literature review outlining fault detection, fault-tolerant and flight control strategies. Chapter 3 details the methodology, including dynamic modeling and the NMPC framework formulation. Chapter 4 presents simulation results validating the proposed control framework in fault recovery and agile maneuvering scenarios. Finally, Chapter 5 concludes the thesis and discusses potential directions for future work.

\chapter{Literature Review}
\label{chap:Literature Review}

Control of unmanned aerial vehicles (UAVs) has undergone significant evolution, transitioning from classical linear controllers toward nonlinear, optimization-based, and learning-driven frameworks. Initial approaches such as PID and LQR controllers provided baseline stability and trajectory tracking for fixed-wing and quadrotor systems. However, as UAVs became more agile and complex, with configurations including tilt-rotors, over-actuated platforms, and morphing architectures, advanced techniques like backstepping, sliding mode, and adaptive control emerged to handle system nonlinearities and environmental disturbances. The introduction of Model Predictive Control (MPC), particularly its nonlinear variant (NMPC), enabled the real-time planning of constrained trajectories, integrating actuator limits, system dynamics, and safety constraints into a single optimization problem. In parallel, data-driven approaches such as RL and neural adaptive control have gained traction, offering flexibility in model uncertainty and fault scenarios.

These developments naturally converged with the increasing need for robustness and resilience, particularly in safety-critical applications. Fault detection and FTC systems became essential in ensuring continued operation despite actuator degradation or sensor failures. Techniques range from sensor-based vibration analysis and model-based estimation to fault-aware control reallocation and learning-augmented compensation strategies.

This literature review presents a structured narrative tracing the evolution from simple passive monitoring techniques to advanced learning-based and adaptive FTC systems, while maintaining the full breadth of technical contributions from diverse approaches.

What follows is an integrated discussion of selected works that illustrate this progression and highlight key insights relevant to the development of a unified NMPC-based control framework for agile and fault-tolerant aerial robots like the M4.



Fault detection in UAVs has traditionally relied on sensor-based methods, and Ghalamchi and Mueller \cite{ghalamchi2018vibration} present a lightweight, sensor-efficient approach using onboard accelerometer data to detect propeller damage via vibration spectrum analysis. Their method leverages the built-in IMU of multicopters and does not require additional sensors or complex modifications. By analyzing discrete Fourier transforms of vibration signals during different flight trajectories, they isolate the frequency associated with a damaged propeller. The novelty lies in the use of distinct flight patterns to cause differential motor speeds, making it possible to identify which propeller is faulty based on the spectral peaks. Experiments using a small quadcopter platform validate the technique for both single and multiple damaged propellers, showing consistent and distinct spectral signatures. This method is particularly useful for low-cost UAVs with limited payload and computational resources, as it supports early-stage damage detection without model knowledge. Despite its simplicity, the approach requires multiple flight maneuvers and is sensitive to ambient vibration noise. Nonetheless, the study provides a practical pathway for trajectory-dependent in-flight fault diagnosis and is a foundational contribution to passive fault monitoring in multicopters.

Building upon such passive methods, Pourpanah et al. \cite{pourpanah2018anomaly} present a two-part fault detection system for UAV motors and propellers using motor current signature analysis (MCSA) and vibration spectrum analysis (VSA). Current signals are captured and analyzed through fast Fourier transform to extract harmonics, which are classified using a fuzzy adaptive resonance theory (ART) neural network. For propeller monitoring, accelerometer data is used to derive statistical features that feed a Q-learning-enhanced Fuzzy ARTMAP (QFAM-GA) neural network. Feature selection is optimized using a genetic algorithm, reducing computational complexity while maintaining classification accuracy. The experimental setup uses Arduino-linked sensors to capture both current and vibration data. The proposed system achieves a classification accuracy of 95.35\% for motor faults and reliably identifies damage scenarios in propellers—categorized into normal, 5\%, 10\%, and 15\% broken states. Comparative results show QFAM-GA outperforms traditional classifiers like SVM, KNN, and Naive Bayes, particularly under noisy input conditions. This hybrid framework demonstrates the potential of combining signal processing and intelligent learning to implement real-time UAV component monitoring. However, the system depends on manual tuning of neural network parameters and extensive training datasets, which may limit generalizability across UAV platforms without retraining.

Transitioning from detection to diagnosis, Palanisamy et al. \cite{palanisamy2022fault} propose a physics-informed model-based technique for propeller fault detection and performance monitoring in electric UAVs. Their work centers on estimating aerodynamic degradation in propellers using only powertrain electrical measurements. A complete powertrain model, integrating the electronic speed controller (ESC), BLDC motor, and propeller, is developed in state-space form. The system is then used to simulate propeller damage conditions under controlled parameter variations. The core contribution is an EKF-based estimator that tracks aerodynamic torque coefficients (Cq) from motor current and voltage measurements to detect off-nominal propeller behavior. Numerical simulations with added Gaussian noise validate the robustness of the EKF estimator under multiple damage levels, 5\%, 10\%, and 50\% reduction in propeller diameter. The approach achieves rapid convergence ($<$0.02s) in estimating both angular velocity and aerodynamic parameters, enabling timely fault identification. However, the methodology assumes availability of high-frequency current sampling from all motor phases, which may not be feasible in smaller UAVs. Despite this, the proposed framework offers a scalable, sensor-less fault detection system rooted in system identification, suitable for incorporation into UAV health-monitoring architectures.

In parallel, researcher began integrating fault diagnosis into control systems. For instance, Mueller and D’Andrea \cite{mueller2014stability} explore aggressive fault-tolerant control in quadcopters capable of stable flight despite the complete loss of up to three propellers. Their work departs from incremental adaptation and focuses on achieving full reconfigurability through an over-actuated design and advanced control allocation. The proposed control strategy leverages a physical model of the quadcopter that predicts dynamic behavior under various actuator loss configurations. They show that by dynamically reallocating thrust vectors and sacrificing control authority over yaw, stable hover and limited maneuvering can still be maintained with only one functioning motor. Experiments validate this result on a custom-built quadrotor with extended arms and redundant motor placements. Their approach is based on the principle of "loss accommodation" rather than fault isolation and recovery, and presents a paradigm shift for multicopter FTC design. Though it imposes constraints on available torque axes, the method proves that complete rotor failures do not necessarily lead to vehicle loss if handled intelligently. This work significantly raises the bar for what is achievable in extreme fault scenarios and serves as a benchmark for physical robustness in multicopter design.

The next wave of research focused on integrating estimation and control, particularly within nonlinear and predictive control frameworks. Nan et al. \cite{nan2022nonlinear} propose a NMPC framework designed for fault-tolerant trajectory tracking in quadrotors. Their architecture uses an adaptive control allocation method that integrates fault estimation into the MPC loop to dynamically adjust input bounds. The controller accounts for varying thrust constraints under actuator faults and maintains feasibility through input relaxation. The novelty of their contribution lies in fusing actuator health estimation with constraint-aware trajectory optimization. They introduce an observer-based estimation layer to track the effective thrust of each rotor, which is then used to adapt the control bounds in real time. Extensive simulations and experimental flight tests demonstrate that the NMPC can track aggressive trajectories even with partial or complete rotor faults. The controller ensures graceful degradation of performance rather than abrupt failure, and avoids infeasibility via slack variables in the optimization formulation. Compared to baseline MPC methods, the proposed solution exhibits higher resilience, maintaining lower tracking error and stability under multiple fault severities. The framework is particularly suited for autonomous quadrotors requiring high reliability, and can be extended to formation flight and collaborative UAV scenarios.Mao et al. \cite{mao2024propeller} extended this by proposing a comprehensive framework for FTC of quadrotors by combining damage estimation and adaptive control. The novelty of the study lies in quantifying the aerodynamic degradation caused by propeller damage and incorporating this estimation into an NMPC controller. Their approach involves a model-based estimation scheme using force-moment data to infer propeller degradation coefficients, followed by adaptive compensation via control reallocation. The method accommodates changes in control effectiveness matrices, ensuring smooth integration with fault-aware NMPC. Simulation and real-world experiments with damaged propellers validate the model’s robustness. Notably, the quadrotor remains stable and can track trajectories even when individual propeller effectiveness is reduced by up to 40\%. The controller dynamically adapts to asymmetric thrust distributions and avoids reliance on external fault identification modules. The approach also includes pre-flight calibration and real-time update mechanisms to minimize mismatch between the identified model and actual UAV dynamics. This tight coupling of fault estimation and control reallocation allows for resilient operations without abrupt behavior changes. The proposed framework is particularly useful in real-world missions where physical inspection of propellers is not feasible, and highlights the importance of coupling estimation and control in UAV fault tolerance.

Simultaneously, vision-based and sensor-light FTC strategies gained traction. Sun et al. \cite{sun2021autonomous} address the challenge of vision-based fault-tolerant flight by enabling autonomous quadrotor navigation despite rotor failure using only onboard vision sensors. The key innovation lies in comparing two visual modalities, traditional frame-based and event-based cameras, for their fault tolerance under aggressive maneuvers. Event cameras, which output pixel-level brightness changes asynchronously, offer high temporal resolution and low latency, making them particularly suitable for high-speed UAV control under degraded conditions. The authors integrate an onboard visual-inertial odometry system with an NMPC controller that compensates for loss of thrust. The system adapts control allocation in real time, leveraging visual feedback for attitude stabilization. The experimental results on a quadrotor with a deliberately failed rotor show that both visual systems can maintain controllability, but event-based vision offers superior responsiveness and lower drift in high-speed scenarios. Their work demonstrates that event-based cameras can act as robust perception backbones in vision-only fault-tolerant flight, particularly where IMU and GPS may be compromised. This study is significant in extending FTC to perception-constrained environments and proves that robust fault adaptation does not necessarily require excessive hardware redundancy if perception is fast and accurate. Ke et al. \cite{ke2023uniform} present a passive FTC framework for quadcopters that can recover stability even under failure of one, two (adjacent or opposite), or three rotors, without requiring any fault information. The core concept is a “uniform” control approach, where the same controller handles all fault scenarios without switching. This is achieved through dynamic control allocation and modeling rotor failures as lumped disturbances in the control input. A modified virtual control structure is introduced to distribute thrust among the remaining actuators. Stability is guaranteed via Lyapunov-based analysis, and the controller’s performance is validated in extensive outdoor experiments. Remarkably, the UAV is able to hover and continue its mission even with three failed rotors, an extreme scenario rarely handled in prior literature. The system avoids reliance on fault detection, which reduces sensor burden and simplifies implementation. The controller’s robustness is attributed to its dynamic disturbance rejection and fault-agnostic design. Source code and flight data are made publicly available, underlining the reproducibility of the work. This research shifts the FTC paradigm toward passive and generalizable architectures that emphasize resilience rather than reactivity, making it particularly relevant for safety-critical UAV applications.

Learning based approach have further enriched the fault-tolerance landscape. O’Connell et al. \cite{oconnell2024learning} propose a learning-based fault-tolerant flight control system for quadrotors that operates effectively under minimal sensing and actuator degradation. The method introduces a model-free, adaptive policy trained via reinforcement learning to provide secondary compensation on top of a nominal controller. The key contribution is a hierarchical control architecture where the inner loop remains stable while the outer loop compensates for control asymmetries introduced by actuator faults. The training is performed in simulation using domain randomization to enable sim-to-real transfer. During flight, the learned policy uses onboard IMU and barometer readings to infer actuator performance and compensate for thrust discrepancies in real time. Experimental results on hardware show that the quadrotor can successfully track aggressive trajectories even under partial motor failure, maintaining hover and maneuverability with as few as two functioning rotors. Unlike most FTC systems, this method does not require prior fault diagnosis, and the policy is robust to noise and model uncertainty. The work is particularly impactful because it reduces hardware complexity and eliminates dependence on external observers or estimators. It demonstrates that with proper training and architecture, learning-based controllers can offer lightweight, plug-and-play solutions for agile fault-tolerant UAV operation. Similarly, Liu et al. \cite{liu2024reinforcement}propose a hybrid fault-tolerant control framework for quadrotor UAVs using reinforcement learning (RL), specifically addressing actuator faults. The method integrates an auxiliary RL-based controller trained using the Proximal Policy Optimization (PPO) algorithm, which outputs compensatory control signals to support a baseline PID controller. This design ensures stability and trajectory tracking even under partial motor failures. The RL controller operates at a lower frequency, using the difference between actual and nominal UAV states to generate fault-compensating actions. Simulation experiments—conducted with varying fault magnitudes and randomized motor failures—demonstrate superior trajectory tracking and disturbance rejection compared to non-fault-tolerant controllers. Real-world validation on a Quanser QDrone shows the system’s effectiveness in compensating for actuator bias faults, maintaining hover stability and reducing tracking errors. Unlike traditional methods that rely on fault detection or model adaptation, this strategy is model-free and data-driven, enabling rapid online compensation without explicit fault classification. The hierarchical architecture separates fast, nominal control from slower, adaptive compensation, minimizing computational burden while ensuring robust performance. Overall, this work highlights the practical feasibility of integrating RL into UAV control stacks, offering fault resilience without sacrificing efficiency or requiring hardware redundancy.

Mechanical design innovations, such as tilt-rotor and over-actuated configurations have also enabled improved FTC capabilities. Nemati et al. \cite{nemati2016stability} explore stability and control of a tilting-rotor quadcopter under propeller failure. The UAV design enables thrust vectoring through motor tilting, which provides an additional control degree of freedom critical for fault tolerance. The authors derive a full 6-DOF nonlinear dynamic model that incorporates tilt dynamics and implement a backstepping-based fault-tolerant controller. The controller reconfigures control authority among functioning rotors and tilting angles when one propeller fails. Simulations and experiments show that even with a rotor out, the UAV maintains altitude, attitude, and trajectory tracking, aided by vectoring thrust using the tilt mechanism. The advantage of the proposed design lies in its mechanical redundancy that complements algorithmic FTC methods. By physically redirecting thrust, the vehicle achieves stabilization without requiring excessive computational resources or observer tuning. However, the added mechanical complexity introduces mass and power consumption trade-offs. Overall, the paper highlights how tilt-rotor mechanisms, when combined with robust control design, can significantly extend fault resilience. This design concept is particularly useful for missions involving confined spaces or hover-intensive tasks, where motor failure without vectoring could result in loss of control. Whereas, Ryll et al. \cite{ryll2012} develop a dynamic model and controller for a tilt-rotor quadcopter, incorporating both translational and tilt-induced rotational dynamics to support fault-tolerant behavior. The UAV design allows each rotor to tilt about an axis, offering an extra degree of control authority when a rotor fails. The proposed model captures the effects of tilt on the net thrust and torque vectors, enabling precise control reallocation during fault conditions. The control strategy uses a nested loop design: an outer loop for position tracking and an inner loop for tilt and attitude control. Simulations under rotor fault scenarios demonstrate that the system can maintain stable flight and follow trajectories by adjusting tilt angles rather than relying solely on rotor thrust. A comparative analysis with a conventional fixed-rotor quadcopter shows that the tilt-rotor configuration significantly improves fault recovery capability. This work underscores the potential of tilt mechanisms not just for agility but for passive redundancy, offering a mechanical pathway to enhance resilience alongside algorithmic controllers. Futhermore, Ji et al. \cite{ruihang2020} develop a dynamic model and control strategy, tailored to these architectures, for a fully actuated tilting quadcopter, where each rotor can adjust its orientation to provide additional control authority. This mechanical configuration enables decoupled control of translation and rotation, enhancing maneuverability and fault resilience. The authors derive a nonlinear 6-DOF model that incorporates gyroscopic effects, aerodynamic damping, and inertia changes from tilting. To ensure rapid and robust tracking, they propose an Adaptive Fast Finite-Time Controller (AFFTC) based on Lyapunov theory. The controller achieves convergence in finite time and avoids chattering issues typically seen in sliding mode control. Simulations compare the proposed controller against traditional finite-time control approaches, showing superior performance in tracking accuracy and robustness to parameter variations and external disturbances. The design includes a two-loop structure, with the inner loop handling attitude and the outer loop managing trajectory tracking. An adaptive estimation law adjusts for modeling uncertainties in real time. The study highlights the advantages of combining mechanical overactuation with fast, adaptive control and is particularly relevant for UAVs operating in constrained or unpredictable environments where fault tolerance and precision are critical.

Control architectures have also diversified. Robust and sliding mode controller remain prominent. Mallavalli and Fekih \cite{mallavalli2019fault} introduce a FTC method for quadrotor UAVs that combines observer-based fault estimation with robust backstepping and sliding mode control. Their framework is designed to mitigate actuator faults without requiring system reconfiguration or redundancy. The approach begins with a fault detection observer that monitors system inputs and outputs to estimate fault magnitude. These estimates are fed into a sliding mode controller that ensures bounded tracking errors despite model uncertainty and actuator degradation. To enhance robustness, a backstepping layer shapes the control signal to compensate for nonlinear dynamics, reducing chattering effects common in sliding mode implementations. Simulation results demonstrate the FTC system can maintain stable flight with one or more degraded actuators, and is effective against both bias-type and loss-of-effectiveness faults. A Lyapunov-based analysis confirms the stability and convergence properties of the closed-loop system. Unlike many FTC methods, this design does not require prior fault isolation or exact actuator fault models, making it suitable for real-time applications where system complexity or sensor limitations restrict extensive diagnosis. The work bridges model-based and robust control methods, contributing to fault-resilient UAV design with minimal computational overhead. Liang et al. \cite{liang2024high} propose a high-accuracy adaptive robust fault-tolerant controller for quadrotors that accounts for actuator uncertainties and aerodynamic drag. The controller is built upon a sliding mode structure augmented with an adaptive law to estimate unknown disturbances and parameter variations. The authors emphasize compensation for aerodynamic effects, often ignored in prior works, which significantly affect quadrotor behavior during aggressive maneuvers. Their method integrates a disturbance observer with parameter adaptation to achieve accurate trajectory tracking despite simultaneous faults and environmental perturbations. A robustness filter is added to suppress chattering caused by sliding mode discontinuities. Simulation results demonstrate that the UAV can track complex trajectories with high precision, even under compound faults and wind disturbances. The study further highlights that the adaptive update law converges rapidly, ensuring bounded estimation errors. Compared to baseline adaptive sliding mode controllers, this design offers superior performance in drag-heavy and fault-prone conditions. The proposed architecture is particularly applicable to outdoor UAV missions involving high-speed flight, where both fault resilience and aerodynamic accuracy are essential for maintaining control integrity.

Meanwhile, neuromorphic computing and neural estimation have emerged. Wei et al. \cite{wei2025spiking} propose a novel attitude tracking controller for UAVs driven by SNNs, targeting scenarios with actuator faults and real-time constraints. SNNs mimic biological neurons, providing ultra-fast inference and low power consumption, which make them ideal for onboard FTC. The controller incorporates a robust adaptive law to estimate fault severity and nonlinear disturbances, while the SNN handles feedback processing and control signal generation. A Lyapunov-based convergence analysis guarantees global stability and fault accommodation. The proposed system is tested in simulation under various fault levels, demonstrating superior tracking accuracy and convergence speed compared to traditional neural network controllers. The spiking-based control architecture is not only computationally efficient but also biologically inspired, offering a promising avenue for neuromorphic flight control. This paper is among the first to apply SNNs in FTC for UAVs, and it opens the door to future research on integrating spiking neuromorphic processors with fault-resilient flight platforms. The approach is particularly relevant for lightweight UAVs where energy efficiency and onboard processing capabilities are limited. While, Abbaspour et al. \cite{abbaspour2020neural} present a neural adaptive fault-tolerant control approach for UAVs, focused on compensating for actuator faults using an online-trained neural network embedded within a backstepping control architecture. The adaptive neural network estimates unknown fault functions and nonlinearities in the UAV system in real time, enabling the controller to adjust its output accordingly. A Lyapunov-based design ensures system stability and convergence of tracking errors. The framework is particularly effective for actuator degradation scenarios such as partial loss of effectiveness, saturation, and bias-type faults. The paper’s novelty lies in its ability to maintain performance without explicit fault isolation or pre-defined failure thresholds. Simulations confirm that the method handles time-varying faults and strong coupling between translational and rotational dynamics. The adaptive neural controller is lightweight enough for onboard implementation, requiring no external observers or diagnostic units. The authors also address practical concerns such as measurement noise and unmodeled dynamics. This work contributes significantly to the development of plug-and-play fault-tolerant controllers that require minimal fault information, relying instead on the learning ability of neural networks to close the loop.

Networked UAV operations and cooperative FTC have also progressed. Miao et al. \cite{miao2024fixed} develop a fixed-time collision-free formation control strategy for multiple UAVs operating under actuator faults. The proposed method ensures that formation tracking errors converge to zero within a predefined time, regardless of initial conditions, a critical advantage in time-sensitive missions such as search and rescue or aerial surveillance. Each UAV in the fleet is equipped with a fault-tolerant controller based on a sliding mode formulation, augmented with barrier Lyapunov functions to avoid inter-agent collisions. The control framework accommodates actuator faults modeled as loss-of-effectiveness and includes online adaptation to handle system uncertainties. Rigorous theoretical analysis confirms that the closed-loop system satisfies both collision avoidance and convergence guarantees. Simulation results with randomly distributed UAVs and injected actuator faults validate the effectiveness of the approach. Notably, the system avoids excessive control effort, thanks to a properly designed triggering mechanism. The study bridges the gap between fault-tolerant control and cooperative formation tracking, offering a scalable solution for multi-agent aerial systems. It sets the groundwork for more complex applications involving fault-resilient coordination in heterogeneous UAV swarms. Yao et al. \cite{yao2024} propose a fault-tolerant control architecture for an overactuated UAV platform composed of quadcopters connected by passive hinges. The overactuated structure provides input redundancy, allowing the vehicle to redistribute control commands across multiple degrees of freedom when actuators fail. The authors introduce a null-space-based control allocation method that optimizes thrust distribution in the presence of actuator degradation. Their controller integrates model-based prediction with onboard feedback to determine the most efficient control action while satisfying physical constraints such as actuator saturation and hinge limitations. Simulations validate the method across various fault scenarios, including full rotor failures and dynamic transitions. The system remains stable and responsive even during aggressive maneuvers. Compared to conventional quadrotor control, the proposed method offers greater fault resilience and more flexibility in trajectory tracking. The control allocation strategy significantly reduces the computational load compared to nonlinear optimization-based methods. This study contributes to the emerging field of overactuated aerial vehicles, showing that with appropriate fault-aware control logic, complex hardware configurations can offer enhanced safety and performance.

Hybrid control approaches, combining model-based and learning-based elements, offer compelling trad-offs. Sohege et al. \cite{sohege2021novel} propose a hybrid FTC approach for UAVs based on robust RL, integrating model-based and data-driven strategies. The controller consists of a base MPC structure enhanced with a robust RL compensator trained using adversarial scenarios. The RL component is trained to handle abrupt system changes and unexpected actuator faults by generating adaptive control signals that supplement the nominal controller. A key contribution is the robustness-aware reward structure during training, which guides the RL agent to avoid unsafe actions while maximizing performance. The system is validated on a quadrotor experiencing actuator degradation and thrust asymmetry, showing that the controller maintains stability and tracking accuracy across a wide range of failure conditions. Compared to standalone RL or MPC methods, the hybrid approach demonstrates faster recovery and reduced control effort. Importantly, the authors design their framework to be computationally efficient, allowing potential onboard deployment. This work contributes significantly to fault-aware autonomy by demonstrating that hybrid architectures can leverage the strengths of both model knowledge and learning to achieve high resilience without compromising control interpretability. Shen et al. \cite{shen2021} present an adaptive learning-based fault-tolerant control approach for quadrotor UAVs subject to time-varying CoG and full-state constraints. The paper addresses a critical gap in FTC literature, handling parameter drift caused by payload changes or fuel consumption, which affect system inertia and dynamics. The proposed method combines barrier Lyapunov functions with neural adaptive estimators to compensate for CoG shifts and actuator faults simultaneously. The controller ensures all state variables remain within safety boundaries and achieves asymptotic tracking despite the presence of significant nonlinearities and uncertainties. Stability is proven using composite Lyapunov functions, and simulations demonstrate the method’s robustness under varying CoG and actuator effectiveness. A key advantage of this work is its generalizability to multi-payload delivery UAVs, which often face mass redistribution in real-world missions. By addressing both structural changes and actuator faults in a unified framework, the authors contribute a resilient and flexible control strategy suited for long-duration or dynamic UAV operations.

Applications have extended into unconventional UAV forms. Hao et al. \cite{hao2022fault} propose a robust position tracking control strategy for tilt tri-rotor UAVs under actuator fault conditions. The tri-rotor configuration, inherently underactuated, poses additional challenges in FTC. The authors design a backstepping-based controller that accounts for the nonlinear dynamics of the tilt mechanism, enabling both rotational and translational stabilization. Their approach incorporates fault estimation into the control loop using a sliding-mode observer to monitor the effectiveness of actuators in real time. The novelty of the method lies in the hierarchical structure: a fast inner-loop manages attitude control, while a slower outer-loop handles trajectory tracking. Experimental validation is carried out using a custom-built tilt tri-rotor platform, showing that the system maintains accurate trajectory tracking despite partial motor loss and tilt actuator limitations. Compared to conventional quadrotors, the tri-rotor system demonstrates higher maneuverability and energy efficiency, making it suitable for urban and indoor environments. The study emphasizes how coupling mechanical flexibility with intelligent control design can improve fault resilience, and it sets a precedent for further research in tilt-based fault-tolerant aerial vehicles. Whereas, Sababha et al. \cite{sababha2015} present a rotor-tilt-free tricopter UAV and analyze its modeling and stability under actuator faults. Unlike traditional tricopters that rely on servo-driven tail rotors for yaw control, this novel design uses differential thrust and torque compensation without mechanical tilt. The authors derive a full 6-DOF nonlinear model capturing aerodynamic coupling, asymmetric thrust distribution, and inertial effects. A PID-based control system is developed and tuned using simulation data to maintain roll, pitch, and yaw stability. The paper explores fault scenarios involving rotor loss and thrust degradation, showing that the tricopter can maintain limited maneuverability through dynamic control redistribution. Though the controller does not include advanced adaptive or learning mechanisms, the mechanical simplicity of the platform offers fault tolerance through redundancy in design rather than computation. The model serves as a foundation for future work in adaptive or model-predictive tricopter FTC and represents a minimalist approach to resilient UAV design. This design is especially useful for low-cost, lightweight platforms used in reconnaissance or indoor operations. Mohamed and Lanzon \cite{kara2012} present a novel tri-rotor UAV with full torque and thrust vectoring capabilities through three independently tilting rotors. Unlike conventional tricopters where only one rotor tilts for yaw control, this design incorporates three tilt mechanisms, enabling six degrees of freedom and complete control authority over translational and rotational motion. Each propeller is mounted on a BLDC motor and attached to a servo-controlled tilting assembly. The system is dynamically modeled using Newton-Euler equations, with a clear decoupling of translational and rotational forces through a structured control allocation matrix. To manage the highly coupled nonlinear dynamics, the authors implement a centralized input-output feedback linearization strategy, followed by an H-infinity loop shaping controller, which ensures robust trajectory tracking and noise attenuation. Simulation results show effective position and attitude tracking from non-zero initial conditions, with actuator outputs remaining within physical constraints. The controller achieves a settling time of ~3 seconds, making it suitable for real-time deployment. Compared to underactuated quadrotors, the proposed tri-rotor offers mechanical redundancy, energy efficiency, and agility. This work significantly contributes to expanding the capabilities of fault-tolerant UAV designs and opens new directions for using fully actuated tricopters in confined or precision-demanding environments. Nava and Pucci \cite{gabriele2023} expand the principles to humanoid aerial robots. They present a comprehensive framework for failure detection and FTC tailored to a jet-powered humanoid aerial robot, iRonCub. This multibody flying platform features four jet turbines mounted on its arms and shoulders, enabling vertical flight with considerable internal dynamics complexity. The proposed FTC strategy combines three key modules: (1) an RPM-based failure detection algorithm that identifies turbine loss through reference-to-measured discrepancy thresholds; (2) an enhanced momentum-based flight controller that adapts to turbine shutdowns by modifying control input bounds; and (3) an offline reference generator that avoids singularities and ensures collision-free postures using nonlinear optimization. Upon fault detection, the robot dynamically reconfigures thrust and joint actuation to maintain flight stability. Simulation results in Gazebo and MATLAB show successful recovery from both arm and back turbine failures, with stable trajectory tracking and reduced joint error when using the optimized reference generator. The work underscores the challenge of FTC in high-DOF flying systems and emphasizes the importance of proactive posture adaptation for robust fault response. The method achieves continuity in control inputs and mitigates instability risks via adaptive weight scheduling and constraint parametrization within a QP framework. This study expands FTC beyond classical UAVs into humanoid jet-propelled robotics.

Adaptive estimation techniques have also matured. Nguyen et al. \cite{nguyen2019} propose a joint actuator fault detection and control scheme for hexacopters using sliding mode observers and a nonlinear Thau estimator. Their framework targets practical fault conditions involving partial actuator loss, common in hexacopter operations due to motor wear or collisions. The proposed system consists of a fault detection unit that estimates actuator health based on residuals from nominal dynamics, and a sliding mode controller that adjusts thrust allocation in response. The authors perform validation using a DJI F550 hexacopter and Pixhawk2 flight controller. Experimental results show the system maintains altitude and stability under both single and dual motor faults. A notable strength of this work is its unified treatment of detection and control, enabling fast response without requiring an external diagnostic layer. Moreover, the estimator handles noise and measurement drift robustly, which is critical for field deployment. The combination of low computational overhead and practical implementation makes this framework suitable for small UAVs in commercial and research applications. The work sets a strong precedent for unified FTC architectures in multi-rotor platforms with redundant actuators. Markus et al. \cite{markus2018} introduce the abrupt-SINDy framework, a method for rapid model recovery in nonlinear dynamical systems subjected to abrupt changes, such as actuator faults. The method extends the SINDy algorithm by incorporating a Lyapunov-based change detection scheme that triggers model updates when prediction divergence exceeds a threshold. Rather than re-learning models from scratch, abrupt-SINDy proposes sparse updates to the existing model by modifying only the necessary terms, significantly reducing retraining time and data requirements. The method is validated on a set of dynamical systems, including chaotic oscillators and mechanical systems with component faults. Results show that the approach can recover accurate reduced-order models in less time and with less data compared to traditional regression-based recovery. In the UAV context, this framework can be employed for online adaptation of flight dynamics under faults or structural changes, where full system identification may be infeasible. While the paper is not UAV-specific, its methodology provides a generalizable tool for integrating fast model adaptation into fault-tolerant controllers, making it highly relevant for model-based FTC architectures operating in unpredictable environments.

Thereafter, fractional-order and fuzzy control strategies provide fine-grained fault adaptation. Yu et al. \cite{ziquan2021} present a fractional-order adaptive fault-tolerant synchronization tracking control scheme for networked fixed-wing UAVs facing both actuator and sensor faults. Their method integrates fractional-order SMC with a recurrent wavelet fuzzy neural network to approximate nonlinearities and fault-induced uncertainties. The recurrent wavelet fuzzy neural network serves as a real-time estimator, updating its parameters through adaptive laws. The control strategy enables each UAV to maintain synchronized tracking within a decentralized communication network while compensating for control effectiveness losses and bias faults in actuators and sensors. Notably, the FO approach allows finer tuning of transient and steady-state performance compared to integer-order designs. Simulation and hardware-in-the-loop experiments confirm the robustness and practicality of the method. This work advances fault-tolerant control in distributed UAV systems and highlights the synergy between neural learning and fractional dynamics. Wang et al. \cite{wang2023} introduce an adaptive fault-tolerant controller for a hybrid canard rotor/wing UAV operating during transition flight, a phase that presents both aerodynamic and actuator modeling challenges. Their control framework integrates adaptive SMC with model parameter estimation to handle actuator faults and system uncertainties in vertical-to-horizontal transitions. The method adapts its control law based on real-time measurements, allowing the UAV to maintain stable flight even under multiple simultaneous faults. A key feature is the incorporation of aerodynamic model compensation that adjusts control effectiveness based on the estimated flight mode. The authors provide formal stability proofs using Lyapunov methods and validate their controller through high-fidelity simulations. Results show that the method successfully tracks desired trajectories during transitions and recovers from actuator faults with minimal performance degradation. The paper addresses a crucial gap in FTC literature, as most prior work focuses on hovering or forward flight but not the hybrid phases in between. This work paves the way for fault-tolerant hybrid UAV designs suitable for delivery, reconnaissance, and urban mobility applications. Similarly, Gao et al. \cite{gao2022} tackle actuator fault-tolerant control for flexible flapping-wing UAVs using a rigid finite element method model to capture the nonlinear coupling between structural deformation and flight dynamics. They propose a finite-time adaptive fuzzy sliding mode controller that accounts for unstructured aerodynamic forces and actuator degradation. The framework ensures convergence of tracking errors within a finite time regardless of initial conditions, a critical feature for UAVs operating in turbulent or confined environments. The authors employ a composite Lyapunov function to prove global stability and validate their controller in simulation across multiple fault scenarios. Their results show significant improvements in robustness, control smoothness, and energy efficiency when compared to conventional SMC methods. This study is one of the few that addresses both structural compliance and actuator faults in flapping-wing aerial vehicles, pushing the boundaries of bio-inspired UAV FTC. It is particularly applicable in surveillance, inspection, and rescue tasks where traditional multirotors may be impractical due to space constraints or stealth requirements.

In addition, Wang et al. \cite{wang2024event} introduce an event-triggered fault-tolerant control framework for quadrotor UAVs that addresses actuator saturation and failure using adaptive fuzzy logic. The proposed system integrates prescribed performance functions with adaptive fuzzy logic and sliding mode control to enforce bounded transient and steady-state responses under fault conditions. A distinctive feature of this framework is the event-triggered mechanism that minimizes control updates, thereby reducing computational load and improving efficiency for embedded systems. The controller adaptively estimates uncertain system dynamics and fault severity using a fuzzy rule base, while guaranteeing that tracking errors remain within a predefined performance envelope. Lyapunov analysis proves the stability and convergence of the system, and simulation results show that the method outperforms continuous control strategies in both precision and resource usage. The scheme maintains performance even under high levels of actuator degradation and external disturbances. This paper represents a significant step toward real-time implementation of FTC systems for resource-constrained UAVs, blending robustness, learning, and low computational complexity. Its generality also suggests compatibility with multi-agent UAV systems and hardware-in-the-loop testing, making it suitable for large-scale aerial deployments.

Building on the theme of rapid recovery and system-level resilience, Mazare et al. \cite{mazare2024robust} propose a robust fault detection and adaptive fixed-time fault-tolerant control framework for quadrotor UAVs. The system comprises two layers: a fault detection module using nonlinear observers and a fixed-time adaptive controller that ensures stabilization regardless of initial conditions. Fixed-time convergence is particularly beneficial in UAV applications where quick recovery is critical. The authors design a Lyapunov-based observer that estimates faults in actuator channels, and use this estimate to dynamically update control inputs in real time. The controller is constructed to be robust against both matched and unmatched uncertainties, including environmental disturbances and structural vibrations. Numerical simulations show that the proposed method can restore tracking accuracy within fixed time bounds even under severe fault conditions. Compared to traditional finite-time controllers, this approach eliminates dependence on initial error magnitude and guarantees convergence within a preset time frame. The study demonstrates that fixed-time design is not only theoretically attractive but also practically viable when combined with adaptive estimation. The work is relevant for UAV operations in constrained environments where rapid recovery is essential, and it opens avenues for integrating time-critical control with embedded fault diagnostics.

While the aforementioned works emphasize system robustness and recovery time, Yu et al. \cite{yu2024fault} explore fault-tolerant control for multirotor aerial transportation systems affected by blade damage. The paper focuses on robust trajectory tracking under rotor asymmetry introduced by partially broken propellers, a condition often overlooked in standard fault models. The proposed controller is based on an adaptive backstepping design that incorporates a damage factor into the dynamic equations of motion. The control law compensates for reduced lift and asymmetric thrust distribution, maintaining payload stability during transport missions. A key feature of their design is its ability to decouple rotor dynamics from external disturbances, ensuring the center of mass remains aligned with the flight trajectory. The authors validate the method through simulation scenarios involving both symmetric and asymmetric blade damage under wind gusts and load variations. Their results show significant improvements in control accuracy and disturbance rejection compared to conventional PID and LQR controllers. The study advances FTC by addressing realistic failure modes in logistics-focused UAV platforms. It is especially relevant as UAVs are increasingly used for delivery applications, where blade damage may occur due to environmental factors or repeated mechanical stress.

While Yu et al. address partial mechanical failures, Ahmadi et al. \cite{ahmadi2023} present an active fault-tolerant control system for quadrotor UAVs based on a nonlinear observer and SMC, validated through hardware-in-the-loop experiments. The proposed observer estimates the impact of motor faults on UAV dynamics, treating faults as disturbances and feeding them directly into a robust sliding mode controller. This dual-layer approach enhances disturbance rejection and eliminates the need for explicit fault classification or reconstruction. A key innovation is the observer’s capability to improve the robustness of the sliding mode controller itself, leading to lower control effort and better transient response. The framework is implemented and tested on a real UAV with motor faults up to 40\% loss of effectiveness. Experimental results show the system maintains trajectory tracking accuracy under varying load and environmental conditions. Unlike many theoretical FTC frameworks, this work emphasizes practical feasibility and real-time performance. The modularity of the observer-controller design also enables easy integration into existing flight control stacks. The results confirm that nonlinear observer-based SMC is not only theoretically sound but also viable in real-time embedded environments.

Real-time monitoring and condition assessment is also the focus of Pourpanah et al. \cite{pourpanah2018anomaly}. They develop a condition monitoring framework for UAV motors and propellers using motor current signature analysis (MCSA) and vibration spectrum analysis (VSA). The system employs three-phase current sensors and a tri-axial accelerometer to gather diagnostic data, which is then processed via feature extraction and classified using neural learning models. For motor anomaly detection, the authors use fuzzy adaptive resonance theory (Fuzzy ART), an unsupervised clustering technique that categorizes frequency-domain harmonics into healthy and faulty groups. For propeller condition analysis, a Q-learning-enhanced fuzzy ARTMAP (QFAM-GA) classifier is used in combination with a genetic algorithm to select the most relevant statistical features. Their experimental results confirm that the hybrid system achieves 95.35\% accuracy for motor faults and high precision in detecting varying degrees of propeller damage (5\%, 10\%, and 15\%). The system outperforms traditional classifiers such as KNN, SVM, and Naive Bayes. The key contribution of this work is its lightweight architecture that enables real-time fault monitoring using embedded computing platforms. It offers a non-invasive, data-driven alternative to physics-based modeling and is ideal for small UAVs operating under variable mission profiles or limited sensing resources.

While condition monitoring enables early fault identification, Eltrabyly et al. \cite{eltrabyly2021} push this further by embedding actuator health estimation directly into the control loop. They address fault-tolerant trajectory tracking for quadcopters experiencing actuator failures using a MPC approach. The novelty of their method lies in its ability to handle up to four faulty actuators simultaneously through constrained optimization. They develop a nonlinear moving horizon estimator (MHE) that continuously updates actuator effectiveness and feeds these estimates into the MPC. The integration of fault estimation and trajectory optimization allows the controller to adapt in real time without the need for explicit fault classification or rule-based switching. The system maintains bounded tracking error across multiple fault cases, and simulation results highlight its robustness under complex, time-varying faults. Compared to traditional gain-scheduled or robust MPC strategies, the proposed method offers better adaptability with reduced computational cost. The paper also addresses implementation feasibility, suggesting a pipeline architecture that enables real-time operation. This work is particularly important for mission-critical applications where actuator faults may cascade, and maintaining flight stability becomes increasingly difficult without adaptive fault modeling.

Exploring different control strategies, Ali et al. \cite{ali2016sensors} propose a fuzzy-based hybrid control algorithm for stabilizing a tri-rotor UAV, addressing the system's nonlinear and underactuated dynamics through a novel integration of multiple control paradigms. The proposed architecture combines a Regulation Pole-Placement Tracking (RST) controller with Model Reference Adaptive Control (MRAC), and adaptively tunes RST controller gains using a Mamdani-type fuzzy logic controller. The tri-rotor UAV under study is equipped with brushless DC motors and exhibits strong coupling between translational and rotational motion, as well as substantial gyroscopic and inertial effects. The hybrid controller is designed to mitigate these dynamics, especially under actuator imbalances and disturbances introduced by torque asymmetry. The controller’s effectiveness is demonstrated through nonlinear 6-DOF simulation, which models the system’s aerodynamic forces and moments in detail. A key innovation of the work lies in the MIT rule-based sensitivity analysis used for real-time tuning of the fuzzy-RST control gains. Simulation results confirm that the fuzzy hybrid controller outperforms a standard adaptive RST controller in terms of transient behavior, convergence speed, and robustness to disturbances. Particularly, it shows zero steady-state error and faster response in both attitude and translational control loops. Stability is guaranteed using Lyapunov-based proofs, and multiple fuzzy if-then rules are constructed to regulate controller output in various error conditions. The study effectively integrates linguistic control reasoning with adaptive pole-placement to address the limitations of rigid linear controllers in complex UAV flight dynamics, offering a practical and computationally feasible solution for real-world tri-rotor stabilization.

Furthermore, the role of mechanical actuation in enhancing fault tolerance is emphasized by Nemati and Kumar \cite{nemati2014ACC} as they present a comprehensive modeling and control framework for a tilting-rotor quadcopter, where each rotor can rotate about its arm axis to provide thrust vectoring. This additional actuation transforms the system into an over-actuated platform with six independent control inputs—four for rotor speeds and two for tilt angles—allowing full control over position and orientation. The authors derive a complete 6-DOF dynamic model incorporating tilt-induced forces and torques and establish relationships between tilt angles and roll/pitch attitudes under hover conditions. They show that hovering at non-zero pitch or roll angles is achievable through symmetric tilt configurations. A PD control strategy is implemented using this model to achieve trajectory tracking and stable orientation control. Simulation results demonstrate successful tracking of 3D trajectories with dynamically adjusted tilt angles and rotor speeds, while maintaining control precision. The model accounts for the changes in rotor thrust requirements due to tilting, and corrects for vertical thrust loss using analytically derived compensation laws. This study bridges the gap between dynamic over-actuation and simple quadrotor designs, offering a scalable approach for agile and fault-resilient flight in cluttered or constrained environments. It serves as a key contribution to the growing field of tilting rotor multirotor UAVs.

Finally, Yoo et al. \cite{yoo2010} present a detailed dynamic modeling and control strategy for two configurations of tri-rotor UAVs: a Single Tri-Rotor and a Coaxial Tri-Rotor. The study addresses the inherent challenge in tri-rotor designs, unbalanced yawing moment due to asymmetric rotor placement, by proposing distinct solutions for each type. The Single Tri-Rotor employs a tilting rear rotor controlled by a servo motor, enabling thrust vectoring to counteract yaw torque and enhance agility. The Coaxial Tri-Rotor configuration instead uses paired counter-rotating rotors on each arm, mechanically canceling reaction torque and improving stability. The authors develop complete 6-DOF nonlinear rigid-body equations for both systems, followed by classical PID-based control laws for altitude, roll, pitch, and yaw. Control allocation strategies are derived to translate collective inputs into motor commands, and simulations validate the controller performance under various flight conditions. Both platforms demonstrate rapid stabilization (1–3 seconds rise and settling time) and effective attitude tracking. The study offers valuable insights into mechanically redundant UAV architectures and highlights the trade-offs between agility and stability in single versus coaxial designs. It lays a strong foundation for future implementation of fault-tolerant and agile tri-rotor aerial vehicles in constrained or high-risk environments.

In sum, the evolution of fault detection and control in UAVs reflects a transition from passive, sensor-based strategies to integrated, adaptive, and learning-driven FTC architectures. Emerging paradigms—tilt-rotor mechanisms, hybrid learning-control systems, and biologically inspired neuromorphic solutions—underscore a collective movement toward resilient, scalable, and agile aerial robotics capable of enduring real-world operational uncertainties.

Despite the breadth of FTC strategies explored in the literature, ranging from fuzzy logic and adaptive estimation to over-actuated and mechanically redundant platforms, few approaches have unified fault recovery and agile maneuvering within a single, generalizable control framework. Most methods either rely on explicit fault detection, require separate modules for estimation and compensation, or sacrifice full control authority (e.g., over yaw) to maintain stability.

This thesis addresses this gap by introducing a unified NMPC framework that simultaneously enables robust fault-tolerant control and agile trajectory tracking. Unlike existing methods that separate failure recovery from nominal control, the proposed NMPC formulation inherently adapts to actuator degradation without prior fault classification or controller switching. By leveraging both thrust vectoring and posture manipulation through joint actuation, the system maintains stability and performance even under partial or complete rotor failures.

This work contributes toward the future of resilient autonomous systems by offering a scalable, adaptive, and versatile control strategy for morphing aerial robots such as M4. The integration of fault tolerance and agility in a single optimization framework marks a significant advancement in the design of high-performance, mission-critical UAV systems.

\chapter{Methodology}
\label{chap :: Methodology}

\section{Flight Dynamics Model}
We model the high-fidelity system and reduced-order model which will serve as the prediction model.
We consider the general state-space form for the flight dynamics of our vehicle:
\begin{equation}
\dot{x} = f(x,u),    
\end{equation}
\noindent where \( x \) is the state vector and \( u \) is the control input. For the high-fidelity model, the system is modeled as a collection of rigid bodies representing the main body and its appendages (three lump of mass considered per each leg). Define the generalized coordinates as
\begin{equation}
q = \begin{bmatrix} p_b, \theta_b, q_a \end{bmatrix}^\top,
\end{equation}
\noindent where \( p_b \in \mathbb{R}^3 \) is the position of the main body, \( \theta_b \in \mathbb{R}^3 \) represents its orientation (e.g., Euler angles), and \( q_a \) collects the joint angles of the appendages. The state vector is given by
\begin{equation}
x = \begin{bmatrix} q,\dot{q} \end{bmatrix}^\top.
\end{equation}

The dynamics are derived using the Euler--Lagrange formulation. Let the Lagrangian be
\begin{equation}
L(q,\dot{q}) = K(q,\dot{q}) - V(q),
\end{equation}
where \( K \) is the total kinetic energy (of both the main body and the appendages) and \( V \) is the gravitational potential energy. The Euler--Lagrange equation is
\begin{equation}
\frac{d}{dt}\left(\frac{\partial L}{\partial \dot{q}}\right) - \frac{\partial L}{\partial q} = Q,
\end{equation}
with \( Q \) representing the generalized forces, which include contributions from joint torques (\(\tau_j\)), thruster forces (\(T_k\)), and reactive moments (\(M_k\)). These inputs are related to the generalized coordinates by a mapping matrix \( B(q) \), so that
\begin{equation}
Q = B(q)\, u, \quad \text{with} \quad u = \begin{bmatrix} T_k,\tau_j\end{bmatrix}^\top.
\end{equation}
%


To capture the aerodynamic damping present in the high-fidelity Simscape model, we augment the Euler–Lagrange dynamics with a generalized drag term~$\tau_{d}$.  
Because translational drag is neglected in the current implementation, $\tau_{d}$ contains only the body-frame rotational damping moments. The resulting equations of motion can be written in the standard second-order form as:

\begin{equation}
M(q)\ddot{q} \;+\; C(q,\dot{q})\dot{q} \;+\; g(q) 
\;+\; \tau_{d}(\dot{q}) 
\;=\; B(q)\,u \;+\; J^{\top}\lambda,
\label{eq:eom_with_drag}
\end{equation}

\begin{equation}
\tau_{d}(\dot{q}) \;=\;
\begin{bmatrix}
\mathbf{0}_{3\times 1} \\[2pt]   
D_{\omega}\,\omega_{b} \\[2pt]   
\mathbf{0}_{n_{a}\times 1}       
\end{bmatrix},
\qquad
D_{\omega} = \operatorname{diag}\!\bigl(\gamma,\,\gamma,\,\gamma \bigr)\;>\;0,
\label{eq:rot_drag}
\end{equation}

where $\omega_{b}\in\mathbb{R}^{3}$ is the body-frame angular-velocity vector
and $\gamma$ is a positive scalar damping coefficients for roll, pitch, and yaw.
The drag vector is expressed in generalized coordinates by zero-padding the translational and appendage rows, so it can be added directly to the left-hand side of~\eqref{eq:eom_with_drag} without modifying the input map $B(q)$.

Defining the state as
\begin{equation}
x = \begin{bmatrix} q, \dot{q} \end{bmatrix}^\top,
\end{equation}
the state-space representation is
\begin{equation}
\dot{x} = \begin{bmatrix}
\dot{q} \\
M(q)^{-1}\Big( B(q)\, u + J^T\,\lambda - C(q,\dot{q})\,\dot{q} - g(q) - \tau_{\mathrm d}(\dot q)\Big)
\end{bmatrix}.
\end{equation}

\subsection{Prediction Model}

The prediction model serves as a reduced-order approximation of the full system dynamics, optimized for computational efficiency within the NMPC framework. To construct this reduced-order model, several simplifications are introduced. First, the intermediate limb segments are assumed to have negligible mass and inertia, allowing us to model the appendages as point masses located at the leg ends. Second, the main body is modeled as a lumped mass system with simplified geometry, approximated as a rectangular cuboid, thereby omitting detailed aerodynamic effects and distributed mass properties. Furthermore, the model neglects coupling effects from structural deformations. These assumptions preserve the dominant translational and rotational dynamics while significantly reducing the complexity of the system, enabling predictive optimization.

The prediction model used for control design is obtain as follows. The main body \((m_b)\) is treated as a 6-DoF rigid body and the appendages are assumed to have point mass \((m_l)\), influencing the mapping of thruster forces. The state vector for the ROM is defined as
\begin{equation}
x_{\text{rom}} = \begin{bmatrix} p_b,\theta_b, q_a, \dot{p}_b,\omega_b, \dot{q}_a \end{bmatrix}^\top,
\end{equation}
where \( p_b \in \mathbb{R}^3 \) is the position of the main body, \( \theta_b \in \mathbb{R}^3 \) represents its orientation (e.g., via Euler angles), \( \dot{p}_b \in \mathbb{R}^3 \) is the linear velocity, and \( \omega_b \in \mathbb{R}^3 \) is the angular velocity.

The translational dynamics are given by Newton's second law:
\begin{equation}
m_{net} \ddot{p}_b = f_{\text{ext}}(q_a,u),
\end{equation}
where \( m_{net} \) \((m_b +4m_l\)) is the total mass of the system and \( f_{\text{ext}}(q_a,u) \) includes the net thruster force (adjusted by the appendage configuration \( q_a \)) and gravity. Although the ROM omits many aero–elastic details, we retain the
\emph{rotational} damping that governs yaw stability. The rotational dynamics follow the Newton--Euler formulation:
\begin{equation}
I_b \dot{\omega}_b + \omega_b \times (I_b \omega_b) = \tau_{\text{ext}}(q_a,u) + \tau_{d}(\omega_b),
\end{equation}
with \( I_b \) the inertia matrix of the main body and \( \tau_{\text{ext}}(q_a,u) \) the net external moment (again, influenced by \( q_a \) and the thruster inputs). Also, 

\[
\tau_{d}(\omega_b)= -\,D_{\omega}\,\omega_b, 
\qquad 
D_{\omega}= \operatorname{diag}(\gamma,\gamma,\gamma)>0,
\]
where $\omega_b\in\mathbb{R}^3$ is the body–frame angular‐velocity vector and
$\gamma$ is empirical roll, pitch, and yaw damping coefficient.

The kinematic relationship between the Euler angle rates and the angular velocity is expressed as
\begin{equation}
\dot{\theta}_b = J(\theta_b)\,\omega_b,
\end{equation}
where \( J(\theta_b) \) is the transformation matrix determined by the Euler angle convention.

Thus, the state-space equations for the ROM are:
\[
\dot{p}_b = \dot{p}_b, \qquad
\dot{\theta}_b = J(\theta_b)\,\omega_b,
\]
\[
\ddot{p}_b = \frac{1}{m_{net}}\Big( f_{\text{thrusters}}(q_a,u) + m_b g \Big),
\]
\[
\dot{\omega}_b = I_b^{-1}\Big( \tau_{\text{thrusters}}(q_a,u) + \tau_{d}(\omega_b) - \omega_b \times (I_b \omega_b) \Big).
\]
In compact form, the ROM state-space model is:
\begin{equation}
\dot{x}_{\text{rom}} = \begin{bmatrix}
\dot{p}_b \\
J(\theta_b)\,\omega_b \\
\dot q_a \\
\frac{1}{m_b}\Big( f_{\text{thrusters}}(q_a,u) + m_b g \Big) \\
I_b^{-1}\Big( \tau_{\text{thrusters}}(q_a,u)  + \tau_{d}(\omega_b) - \omega_b \times (I_b \omega_b) \Big) \\
J_a \, u
\end{bmatrix},
\end{equation}
where $J_a$ maps $u$ into the joint acceleration.

\section{Simscape Multibody Model}

To validate the reduced-order prediction model and to rigorously test the NMPC controller design, a comprehensive high-fidelity simulation of the M4 aerial robot was developed using Simscape Multibody, an advanced physics-based modeling environment within MATLAB Simulink. The Simscape model captures the intricate full-body dynamics of the M4 robot, including rigid-body interactions, joint actuation limits, aerodynamic disturbances, and collision dynamics, thereby serving as a realistic digital twin of the physical system.

\begin{figure}[!htbp]
    \centering
    \includegraphics[width=\textwidth]{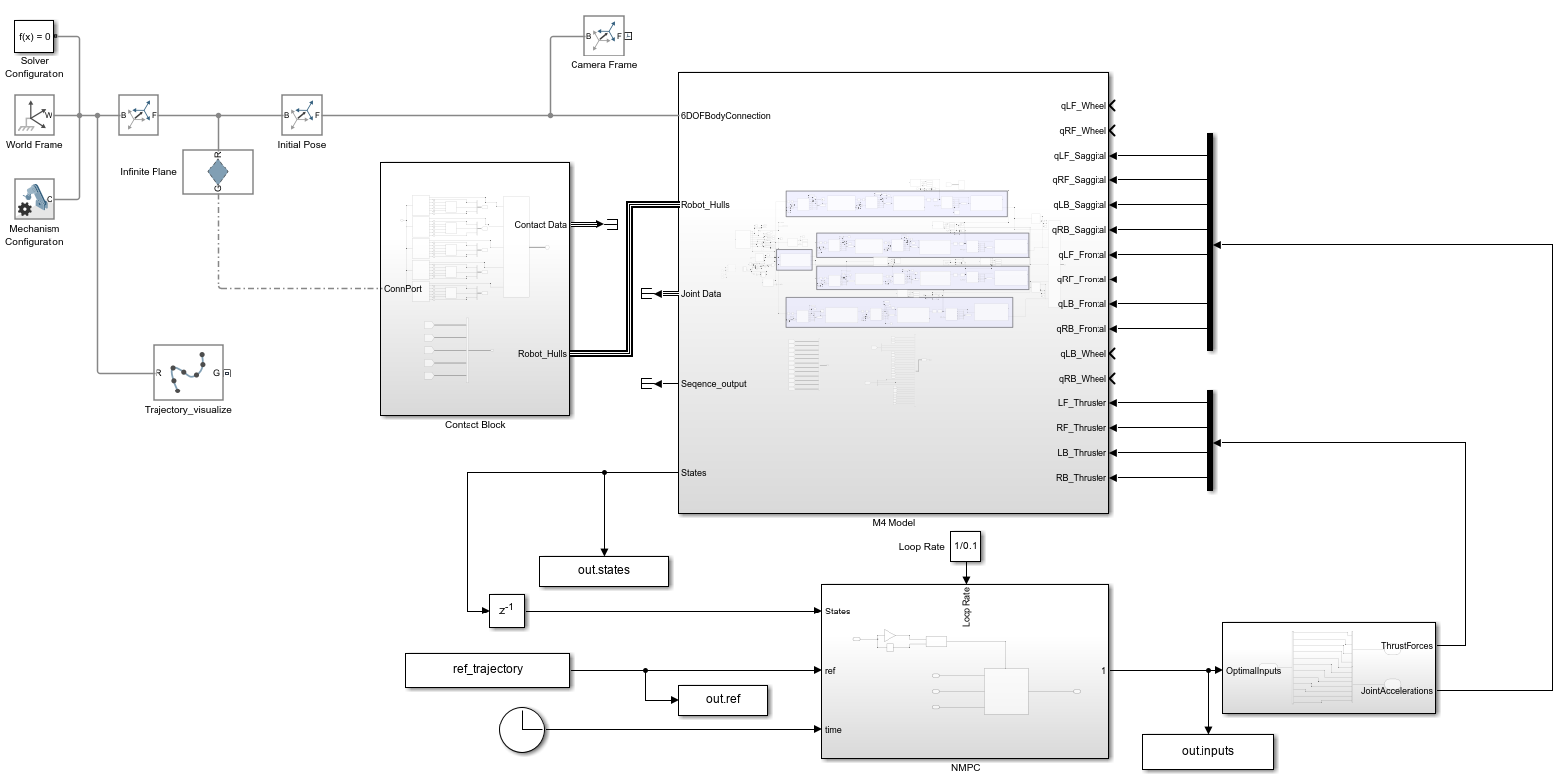}
    \caption{Block diagram of the integrated control and simulation environment, highlighting the feedback loop between the NMPC optimizer and the Simscape multibody plant.}
    \label{fig:simscape_block}
\end{figure}

The model architecture consists of the central body and four articulated appendages, with each limb featuring two rotational degrees of freedom: a sagittal (fore-aft) joint and a frontal (side-to-side) joint. The central body is attached to Simscape's 6-DOF joint block, which allows unconstrained translational and rotational motion in all three spatial dimensions, effectively assigning complete six-degree-of-freedom dynamics to the robot's main body. This block thus accurately captures the robot’s overall translational and rotational dynamics, crucial for realistic simulation of flight behaviors. Each limb joint is modeled using Simscape's revolute joint blocks, and custom STL files corresponding to the individual limb segments are imported to precisely represent the robot’s physical geometry. These revolute joints are constrained to operate within a predefined range of $[0^\circ, 90^\circ]$, ensuring the simulated limb movements closely match the robot's physical joint limits and maintain realistic motion envelopes.

Each limb's geometric parameters, such as mass, inertia, and center of mass locations, were accurately defined in accordance with the physical prototype's specifications, ensuring realistic dynamic behavior. The inertia properties and spatial relationships between limb components are critical factors influencing the robot's overall stability, agility, and response during flight maneuvers. These parameters were fine-tuned iteratively to reflect physical test data, thereby enhancing simulation fidelity.

For actuation, thrust forces and moments are applied along the z-direction at the base of each wheel using external force and torque blocks, where thrust-induced moments are modeled by applying a scaled component of thrust (gain factor $k$) to simulate and aerodynamic effects.

\begin{figure}[!htbp]
    \centering
    \includegraphics[width=1\textwidth]{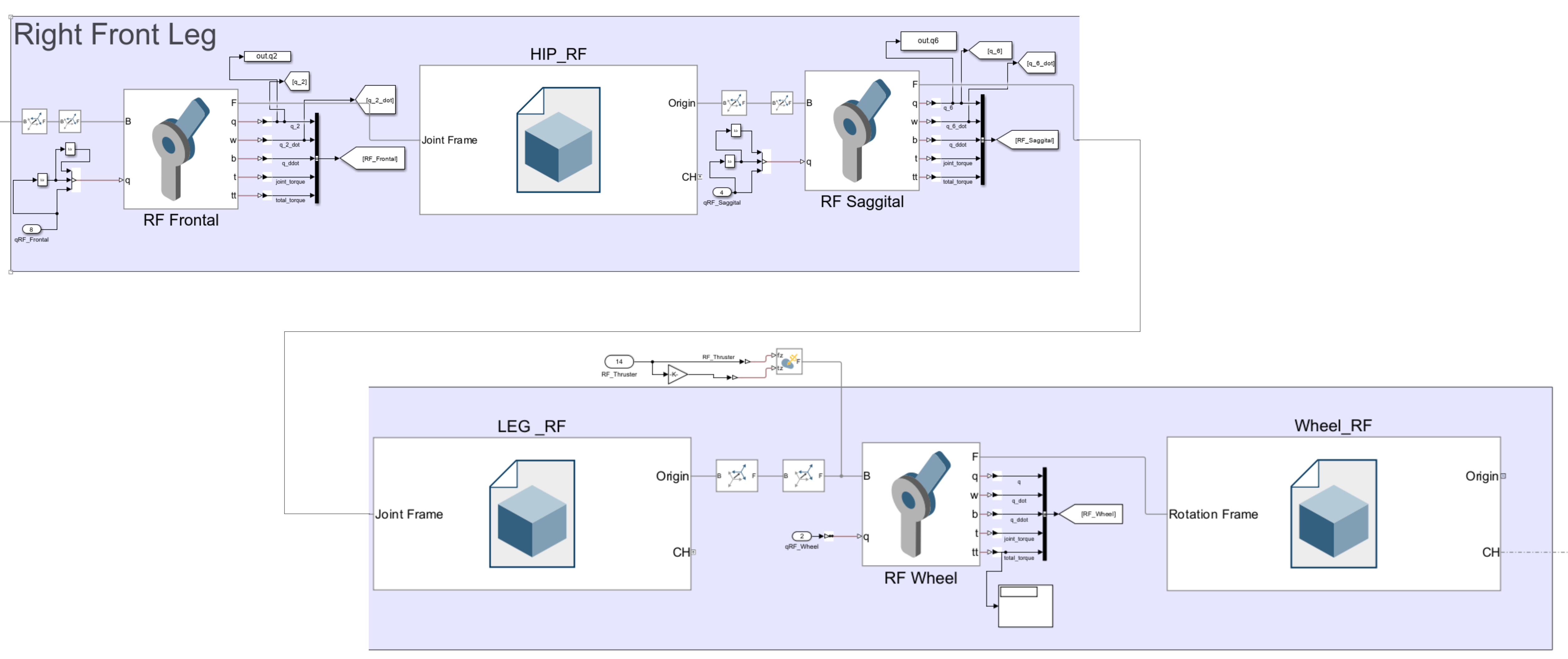}
    \caption{Detailed schematic of leg articulation subsystem: sagittal and frontal joints per limb, STL parts assembly, and placement of thrust and external force blocks. Contact forces are applied between adjacent wheels to enforce physical realism and avoid limb interference.}
    \label{fig:leg_articulation}
\end{figure}

Robot states such as body position, orientation, linear velocities, and angular rates are extracted via a transform sensor attached to the central body frame. Additionally, sagittal and frontal joint angles along with their angular rates are read directly from the revolute joint blocks. These signals are subsequently fed back into the NMPC optimization block to facilitate closed-loop control.

To accurately capture physical feasibility, contact forces are implemented using Simscape’s Spatial Contact Force blocks. Convex hull geometries are defined for adjacent wheels, and contact forces are activated when limbs potentially overlap within the articulation envelope, enforcing non-collision constraints during dynamic reconfiguration.

Aerodynamic effects such as drag and thrust-induced moments are also incorporated for higher-fidelity flight dynamics. Drag is modeled by extracting the body’s angular velocity and applying damping torques proportional to this velocity. The drag forces are applied using an External Force and Torque block with coefficients defined by a gain factor $\gamma$. This introduces realistic roll, pitch, and yaw damping during flight.

\begin{figure}[!htbp]
    \centering
    \includegraphics[width=\textwidth]{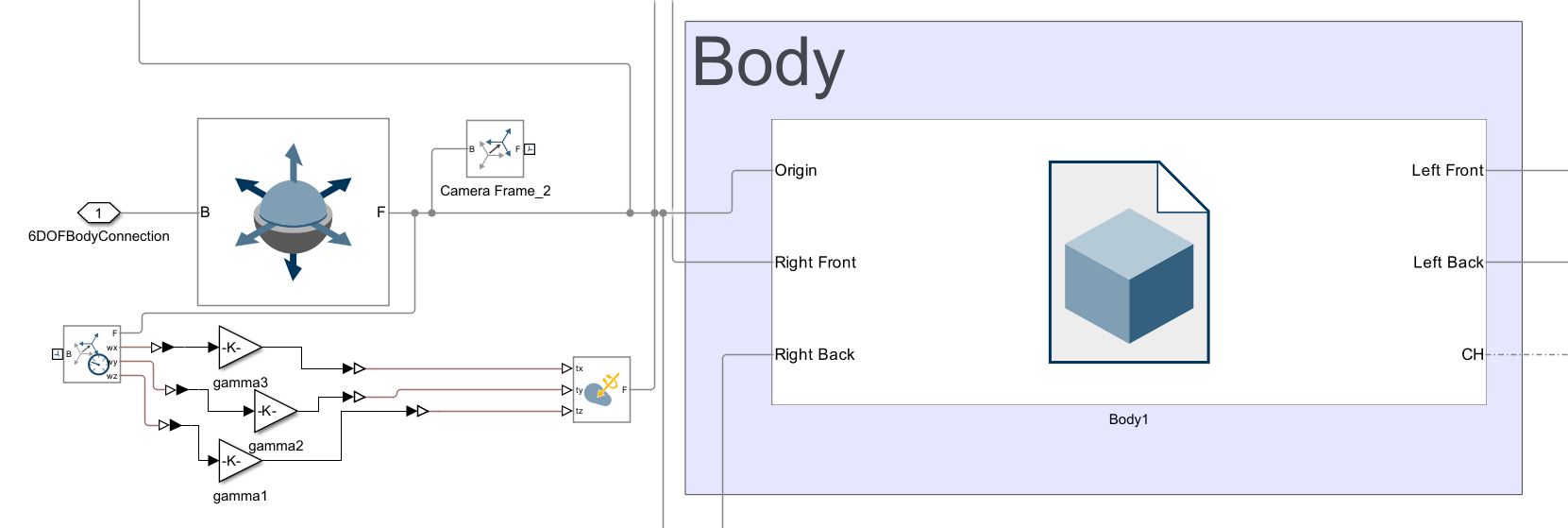}
    \caption{Implementation of aerodynamic drag and thrust-induced moment simulation within the Simulink environment. Drag torques are applied proportionally to angular velocities.}
    \label{fig:body drag}
\end{figure}

The control architecture interfaces seamlessly with the Simscape model via an S-function block, bridging MATLAB's CasADi-based NMPC framework with the Simulink environment. This S-function receives updated body states and reference trajectories from MATLAB workspace variables and initiates computation of optimal NMPC control actions at a discrete sampling interval of $\Delta t = 0.1$~s. This specific interval was selected to optimize performance and computational efficiency during simulation.

The NMPC-generated control inputs, consisting of thrust forces and joint accelerations, are directly applied to the multibody model. Thrust forces $F_{\text{thrust}}$ are implemented through Simscape's External Force and Torque blocks, applied directly at their respective robot reference frames. The joint acceleration signals $\ddot{\theta}_{j}$ computed by NMPC are provided to a Simulink-PS Converter, explicitly configured to interpret these inputs as second derivatives of joint angles. Joint velocities and joint angles are sequentially derived via integration as follows:
\begin{align}
    \dot{\theta}_{j}(t) &= \int \ddot{\theta}_{j}(t)\, dt, \\
    \theta_{j}(t) &= \int \dot{\theta}_{j}(t)\, dt.
\end{align}
Explicit integrator blocks in Simulink recover the joint velocity ($\dot{\theta}_{j}$) from the acceleration signal and subsequently integrate again to obtain the joint position ($\theta_{j}$). These joint angle, velocity, and acceleration outputs from the Simulink-PS Converter are then directly connected to the motion actuation inputs of the revolute joints, ensuring accurate realization of the NMPC-defined joint trajectories. This configuration enables precise posture manipulation and effective thrust vectoring during dynamic flight maneuvers.

\begin{figure}[!htbp]
    \centering
    \includegraphics[width=0.7\textwidth]{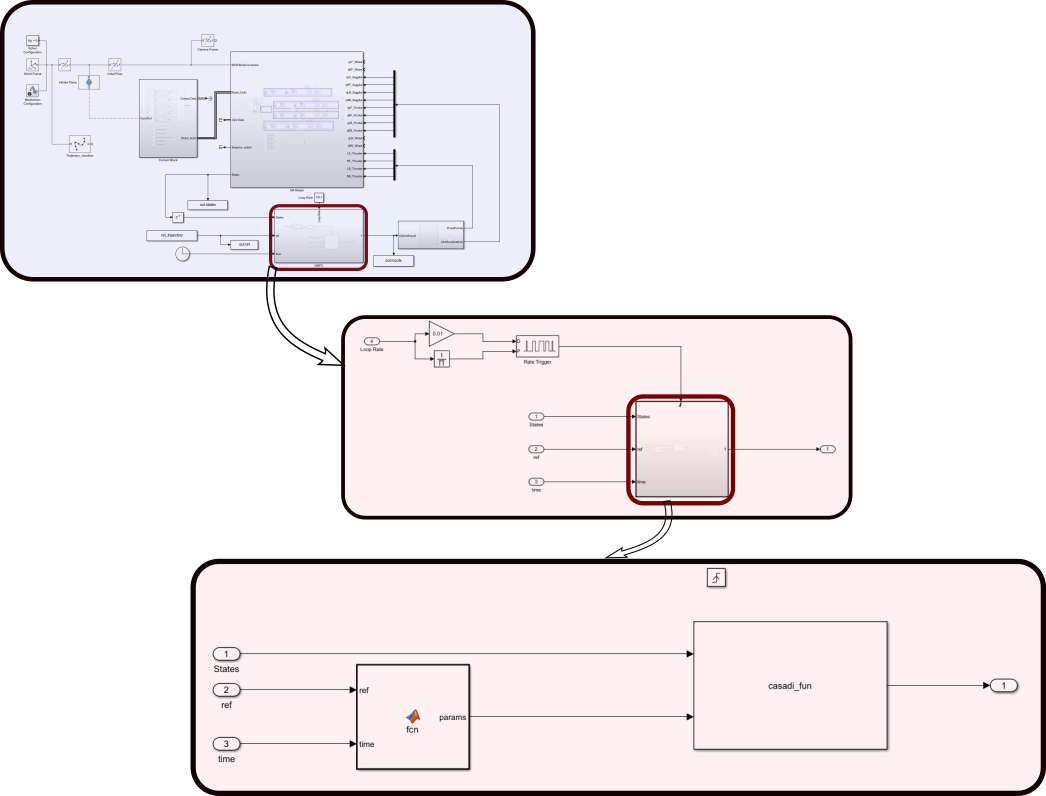}
    \caption{Overview of the control subsystem architecture: The NMPC block processes delayed body state feedback and reference trajectory inputs to compute optimized control actions at a 10Hz update rate.}
    \label{fig:control_subsystems}
\end{figure}

Despite the primarily aerial operation of the M4 robot in this thesis, ground-contact modeling is critical for accurately capturing scenarios involving ground proximity, landing phases, or analyzing complex internal dynamic behaviors. A dedicated contact subsystem is therefore included in the Simscape Multibody model, wherein each limb’s convex hull geometry can establish contact with a theoretically infinite planar ground surface. 

To accurately simulate realistic ground interactions, contact forces are computed considering parameters such as material stiffness, damping coefficients, and frictional properties. Each contacting geometry includes a dedicated contact frame precisely positioned at the instantaneous point of contact. These contact frames define the local contact plane through their xy-plane alignment, while the z-axis orientation serves as an outward normal for the base geometry and an inward normal for the follower geometry. As continuous contact occurs, these frames dynamically shift along the interacting geometries, effectively tracking the instantaneous contact points.

Contact forces applied by the subsystem adhere strictly to Newton’s Third Law, comprising two primary components: the normal force, \( f_n \), and the frictional force, \( f_f \). The normal force acts along the z-axis of the contact frame to reduce penetration depth between geometries. It is computed using a Smooth Spring-Damper model characterized by a spring stiffness of \(1 \times 10^{-4}\, \text{N/m}\), damping coefficient of \(1 \times 10^{3}\, \text{Ns/m}\), and a transition width of \(1 \times 10^{-3}\, \text{m}\). The transition width parameter determines the smoothness of the force application, allowing sharper or smoother transitions based on the selected value. Mathematically, the normal force is expressed as:
\begin{equation}
    f_n = s(d, w)\cdot(k \cdot d + b \cdot d'),
\end{equation}
where \( d \) represents the penetration depth between geometries, \( d' \) is its time derivative, \( k \) denotes the stiffness coefficient, \( b \) denotes the damping coefficient, \( w \) is the transition region width, and \( s(d,w) \) represents the smoothing function ensuring continuous force transitions.

The frictional forces, critical for realistically modeling ground interaction dynamics, are computed using a Smooth Stick-Slip friction model. This model uses a static friction coefficient of \(0.7\), dynamic friction coefficient of \(0.5\), and critical velocity of \(1 \times 10^{-3}\, \text{m/s}\). The resulting frictional force magnitude is given by:
\begin{equation}
    |f_f| = \mu \cdot |f_n|,
\end{equation}
where \( \mu \) is the effective friction coefficient, derived based on the relative motion conditions at the contact interface.

This high-fidelity Simscape model forms the principal validation platform for assessing the performance of the NMPC in both fault-tolerant recovery and agile trajectory tracking. Simulation results derived from this integrated simulation-control environment are systematically presented in the Results chapter.

\begin{figure}[H]
    \centering
    \includegraphics[width=\textwidth]{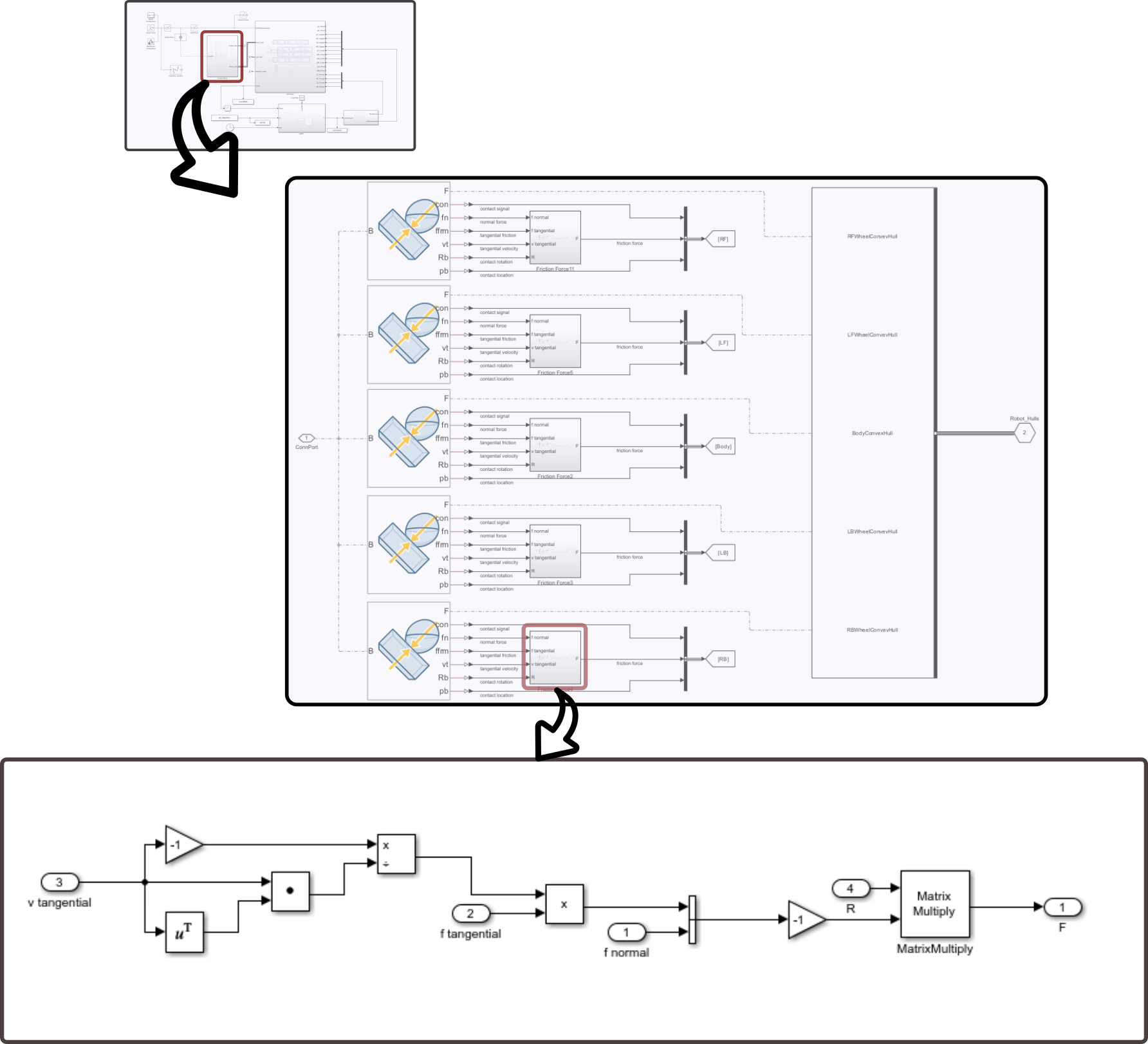}
    \caption{Detailed schematic of the ground contact modeling subsystem in Simscape. Contact force blocks link the convex hull geometries of limbs to an infinite planar ground model.}
    \label{fig:contact block}
\end{figure}

\section{Nonlinear Model Predictive Control}

A unified nonlinear model predictive control framework is developed to enable both fault-tolerant recovery and agile flight for the M4 aerial robot. The controller uses a reduced-order dynamic model as its prediction model and solves a constrained optimization problem at each time step. While the core structure remains identical across modes, the formulation is parameterized differently to meet the requirements of each task. This includes adjustments to thrust limits, cost function weights, and reference trajectory handling. The following subsections outline the distinct configurations used for fault-tolerant and agile flight modes.

\subsection{NMPC for Fault-Tolerant Control}

In fault-tolerant mode, the NMPC framework is designed to prioritize recovery and stabilization following partial or complete loss of actuator effectiveness. The optimization problem is formulated as:

\begin{equation}
\min_{\{u_j\}} \sum_{j=0}^{N_h - 1} \left[ (x_j - x_{\text{ref},j})^\top Q_{\text{fault}} (x_j - x_{\text{ref},j}) + u_j^\top R_{\text{fault}} u_j \right]
\end{equation}

subject to:
\begin{align}
x_{j+1} &= \Phi(x_j, u_j), \quad x_0 = x(t_0) \\
x_{\min} &\leq x_j \leq x_{\max}, \quad u_{\min} \leq u_j \leq u_{\max}
\end{align}

The function $\Phi(x_j, u_j)$ represents the discretized evolution of the reduced-order model (ROM) over one time step. The continuous-time dynamics of the prediction model are given by:

\begin{equation}
\dot{x}_{\text{rom}} = f_{\text{rom}}(x_{\text{rom}}, u)
\tag{3.16}
\end{equation}

and are integrated numerically using a fourth-order Runge--Kutta scheme.

Key characteristics of this configuration include:
\begin{itemize}
    \item \textbf{Thrust Limits:} Thrust inputs are bounded within [0, 30]~N to reflect nominal actuator capacity and ensure realistic saturation behavior during fault conditions.
    \item \textbf{Cost Function:} The positive definite matrices $Q_{\text{fault}}$ and $R_{\text{fault}}$ are tuned to penalize deviations from a hovering or recovery configuration, prioritizing roll and pitch stability over aggressive tracking.
    \item \textbf{Yaw Handling:} In sagittal-only actuation, yaw is unconstrained to allow free spin; in the fully actuated configuration, yaw deviations are penalized to encourage re-stabilization.
    \item \textbf{Control Response:} The controller reallocates thrust and joint torques dynamically upon failure without relying on fault detection modules or switching logic.
\end{itemize}

This configuration ensures that the robot can maintain controlled flight even under severe actuator degradation, leveraging joint actuation for fault compensation.

\subsection{NMPC for Agile Trajectory Tracking}

In agile mode, the controller is tuned to enable fast and precise tracking of aggressive flight trajectories, including sharp turns and rapid acceleration. While the structure of the optimization problem remains the same, two critical differences are introduced: expanded actuation limits and a collocation-based reference generation scheme.

\begin{equation}
\min_{\{u_j\}} \sum_{j=0}^{N_h - 1} \left[ (x_j - x_{\text{ref},j})^\top Q_{\text{agile}} (x_j - x_{\text{ref},j}) + u_j^\top R_{\text{agile}} u_j \right]
\end{equation}

subject to the same dynamics and constraints as before.

As in the fault-tolerant mode, the prediction model dynamics follow:

\begin{equation}
\dot{x}_{\text{rom}} = f_{\text{rom}}(x_{\text{rom}}, u)
\tag{3.16}
\end{equation}

which is integrated over the prediction horizon using a fixed-step Runge--Kutta method.

Key differences in configuration include:
\begin{itemize}
    \item \textbf{Thrust Limits:} Thrust inputs are allowed to vary in a wider range of [0, 50]~N, enabling more aggressive force application to support rapid motion and sharp trajectory corrections.
    \item \textbf{Cost Function:} The positive definite matrices $Q_{\text{agile}}$ and $R_{\text{agile}}$ are selected to place greater emphasis on tracking precision and velocity control, while allowing more joint movement and thrust variation.
    \item \textbf{Reference Handling via Collocation:} Rather than directly using a final reference state at each horizon, a collocation method is used to interpolate intermediate reference states from the current state toward the goal. These intermediate references are updated incrementally over the prediction horizon, ensuring smoother transitions and improved feasibility in high-speed flight.
    \item \textbf{Yaw Behavior:} Yaw error is penalized in the cost function to keep the vehicle’s heading steady during aggressive turns, ensuring the robot points in the same direction even while executing tight maneuvers.
\end{itemize}

By combining expanded input authority with smooth, staged reference tracking via collocation, this NMPC setup enables M4 to execute agile trajectories, including sharp turns at speeds exceeding 10~m/s, while maintaining robust flight performance.


\chapter{Results}
\label{chap:results}

Simulation were performed in Simscape/Simulink, which provides a high-fidelity rigid-body model of the M4 platform.  To get some idea about the robot specifications (weight, thrust force at hover, etc.), some model characteristics of the simulated robot are presented in Table~\ref{tab:model_specs}.  These values also serve as the baseline for normalising control weights and defining thrust-to-weight ratios in the scenarios that follow.

\begin{table}[htbp]
    \centering
    \caption{Design specifications of the simulated M4 robot.}
    \label{tab:model_specs}
    \begin{tabular}{@{}ll@{}}
        \toprule
        \textbf{Property}           & \textbf{Value} \\ \midrule
        Body mass                   & 4.4 kg \\
        Mass of legs                & 0.4 kg \\
        Total mass of robot         & 6 kg \\
        Force due to gravity        & 58.86 N \\
        Nominal thrust per propeller & 14.715 N \\ \bottomrule
    \end{tabular}
\end{table}

At every control cycle the NMPC:

\begin{enumerate}
    \item receives the current state from the Simscape plant,
    \item solves a finite-horizon optimal control problem with fixed weighting matrices, and
    \item returns the optimal joint trajectories and thrusts to the robot model.
\end{enumerate}

These three steps close the loop between the optimizer and the high-fidelity plant, allowing realistic evaluation of recovery transients and aggressive maneuvers.

The Simscape plant was integrated with a fixed step of $10^{-4}\,\mathrm{s}$ using a fourth-order Runge–Kutta (RK4) scheme.  
The NMPC optimizer ran every $0.1\,\mathrm{s}$ with a five-step prediction horizon, holding the control input constant within each interval.  
Table~\ref{tab:sim_params} summarizes the principal simulation parameters.  
The rotor moment–thrust coefficient $k$ and the aerodynamic yaw-damping coefficient $\gamma$ were selected from quantitative analysis on comparably sized aerial platforms. Prior to hardware deployment they would be refined through dedicated system-identification tests on the physical robot.

\begin{table}[htbp]
    \centering
    \caption{Simulation parameters for the high-fidelity Simscape/Simulink simulations.}
    \label{tab:sim_params}
    \begin{tabular}{@{}lll@{}}
        \toprule
        \textbf{Parameter} & \textbf{Description}                 & \textbf{Value} \\ \midrule
        Horizon                     & NMPC prediction length                    & 5 \\
        Simulation step time        & Plant integration step                    & $10^{-4}\,\mathrm{s}$ \\
        Integration scheme          & Plant integrator                          & Runge–Kutta 4\textsuperscript{th} order (RK4) \\
        Controller loop time        & NMPC update period                        & $0.1\,\mathrm{s}$ \\
        Solver                      & NLP solver used inside NMPC               & IPOPT (CasADi) \\[2pt]
        Gain $k$                    & Rotor moment–thrust coefficient $(\tau_i = k\,T_i)$ & 0.055 \\
        Gain $\gamma$               & Aerodynamic damping coefficient           & 0.275 \\ \bottomrule
    \end{tabular}
\end{table}

The NMPC imposed primary state constraints on body orientation and joint angles.  Specifically, roll and pitch are constrained to $[-90^\circ, 90^\circ]$.  
Hip–sagittal joints lie in $[0^\circ, 90^\circ]$ (fully retracted to fully extended) and hip–frontal joints in $[0^\circ, 90^\circ]$ (supinated to pronated).  
Thrust limits are \(\smash{T_i \in [0,\,30]\,\text{N}}\) in fault-tolerant tests (\(\approx 2{:}1\) thrust-to-weight) and
\(\smash{T_i \in [0,\,50]\,\text{N}}\) (\(\approx 3{:}1\)) in agile-flight tests. Joint accelerations satisfy \(\ddot q_i \in [-50,\,50]\,\text{rad/s}^2\).  No explicit bounds are placed on body-frame velocities or joint rates.  These limits reflect the mechanical envelope of the hardware design and ensure that simulated control actions remain physically feasible.

To validate the controller's capability, various simulations were conducted for fault-tolerant and agile-flight scenarios, results of which are presented in the subsequent sections, after a brief discussion on prediction-model fidelity.  The test matrix includes the Stage~1 and Stage~2 fault-recovery cases (instant and progressive single-rotor failures) as well as high-speed turning manoeuvres up to $120^\circ$; together they demonstrate the unified NMPC’s ability to handle both extreme actuator loss and aggressive trajectory tracking without mode-switching or re-tuning.

\section{Prediction Model Performance} 

\begin{figure}[H]
    \centering
    \includegraphics[width=0.8\linewidth]{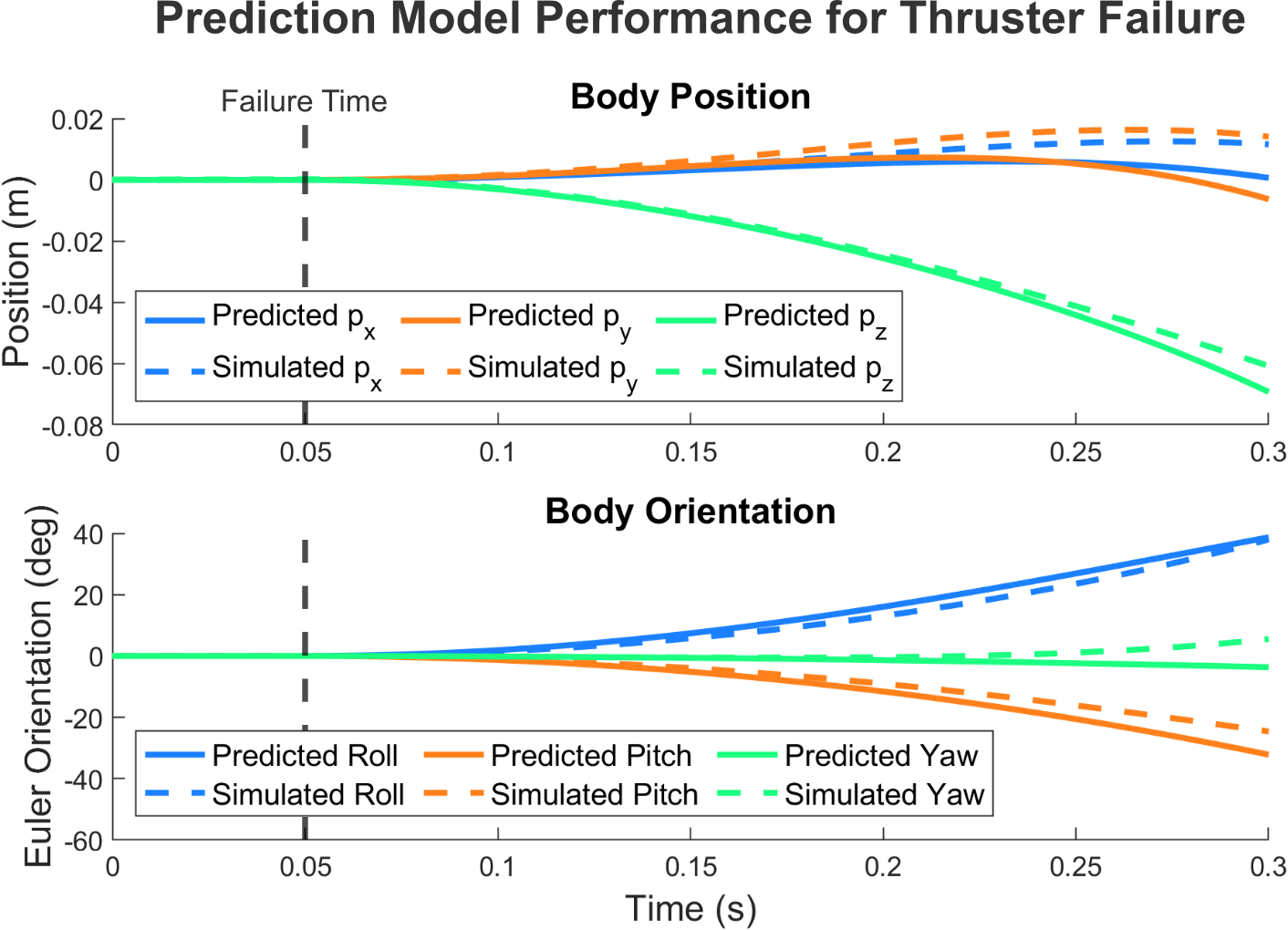}
    \caption{Plot showing simulated states between the NMPC prediction model states and Simscape model during thruster failure event. The predicted model closely resembles the full fidelity model.}
    \label{fig::pmp}
\end{figure}

Figure \ref{fig::pmp} illustrates the performance of the NMPC-based prediction model during an actuator failure scenario. The plot compares the predicted states from the NMPC  reduced-order   prediction model with the actual states obtained from the high-fidelity Simscape simulation of the M4 platform before and after the failure of rotor 4. The robot is in stable hovering condition for the initial 0.2 s. Rotor 4 is allowed to switch off at 0.2 s and the position and Euler angles after this instant are captured in the presented figure. Despite the complexity introduced by the actuator failure, the prediction model closely tracks the states of the full-fidelity model within the range of about 0.3 s. This indicates that the reduced-order model used for prediction effectively captures the essential dynamics required for fault recovery. The high degree of similarity between the predicted and actual trajectories validates the accuracy and reliability of the proposed controller during fault conditions.

\section{Fault-Tolerant Flight}
Fault-tolerance performance was evaluated through two simulation cases for two different models. In the first model, the controller was permitted to actuate only the hip-sagittal joints, while the hip-frontal joints were held in a fully supinated position. In the second model, both hip-sagittal and hip-frontal joints were actuated. Results show that allowing both sets of joints to actuate enabled full stabilization of the robot, particularly in terms of yaw dynamics, during a thruster failure.

\begin{itemize}[leftmargin=1.4em]
\item \textbf{Stage~1 : 1-DoF articulation (sagittal joints only)}
      \begin{itemize}
      \item \textit{Case 1:} forward flight $\;\rightarrow\;$ instant single-thruster failure $\;\rightarrow\;$ recovery to hover
      \item \textit{Case 2:} waypoint-tracking trajectory $\;\rightarrow\;$ progressive single-thruster failure $\;\rightarrow\;$ recover and continue tracking to landing
      \end{itemize}

\item \textbf{Stage~2 : 2-DoF articulation (sagittal $+$ frontal joints)}
      \begin{itemize}
      \item \textit{Case 1:} hover $\;\rightarrow\;$ progressive single-thruster failure $\;\rightarrow\;$ recovery to hover
      \item \textit{Case 2:} waypoint-tracking trajectory $\;\rightarrow\;$ progressive single-thruster failure $\;\rightarrow\;$ recover and continue tracking to landing
      \end{itemize}
\end{itemize}

Importantly, the controller operates without any prior knowledge of fault conditions, eliminating the need for explicit fault detection or controller switching. The same NMPC framework is responsible for both stable flight and fault-tolerant recovery. In a real-world scenario, such faults would likely be detected by onboard sensors due to persistent thrust saturation. The results demonstrate the proposed controller’s robustness and adaptability, enabling recovery from critical actuator failures.

\subsection{Simulation Results for Sagittal-Actuated Joint Model}
The sagittal-actuated joint model corresponds to a system configuration in which each leg of the M4 robot is held in a supinated position with respect to the frontal joint motion, i.e., the frontal joints remain fixed. In this configuration, the NMPC framework enforces an additional constraint to prevent wheel collisions: the combined joint angles on each side of the robot are limited to approximately $110^\circ$. Lastly, no explicit constraint or cost was applied to yaw in the sagittal-only configuration, allowing free rotation about the yaw axis during thruster failure, while all other state constraints remained active.

The simulation results for this model are analyzed in two distinct scenarios: (1) fault recovery during forward flight, and (2) fault recovery during trajectory tracking. The following sections present the performance of the controller in both scenarios and demonstrate the system’s ability to stabilize and recover from actuator failures under the imposed constraints.

\subsubsection{Scenario 1: Fault Recovery During Forward Flight to Hover}


In this scenario, we simulate a perturbation-based fault case during forward flight, where the robot is initially commanded to follow a trajectory in the $x$-direction to reach a goal position. To test the robustness of the NMPC controller, we introduce a failure in thruster 4 at randomized time instances between 3.825\,s and 3.925\,s, while the controller reacts at exactly 4\,s. The objective is to assess the robot’s ability to recover and maintain a hover-like state despite the temporal mismatch between failure occurrence and control response.

\begin{figure}[!htbp]
    \centering
    \includegraphics[width=0.9\linewidth]{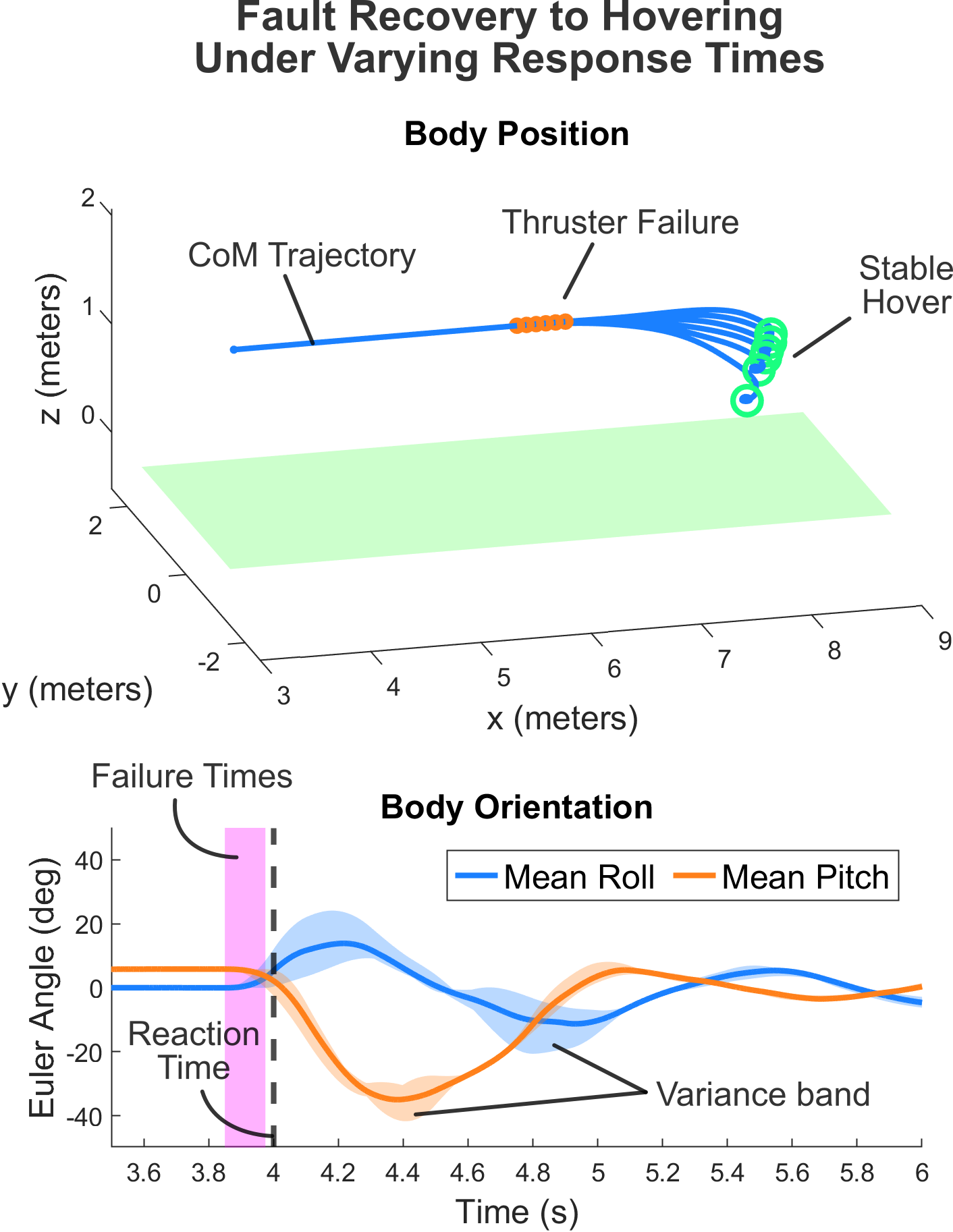}
    \caption{Plots showing the center of mass trajectory and Euler angles of the robot during failure events until it reaches a stable recovery state. A range of response times are tested, with failure events occurring between 3.825 and 3.925s, and the controller reacting to the failure at 4 seconds. The mean roll and pitch are shown with the variance across the various test cases.}
    \label{fig::sa-sc-11}
\end{figure}

\begin{figure}[!htbp]
    \centering
    \includegraphics[width=0.7\linewidth]{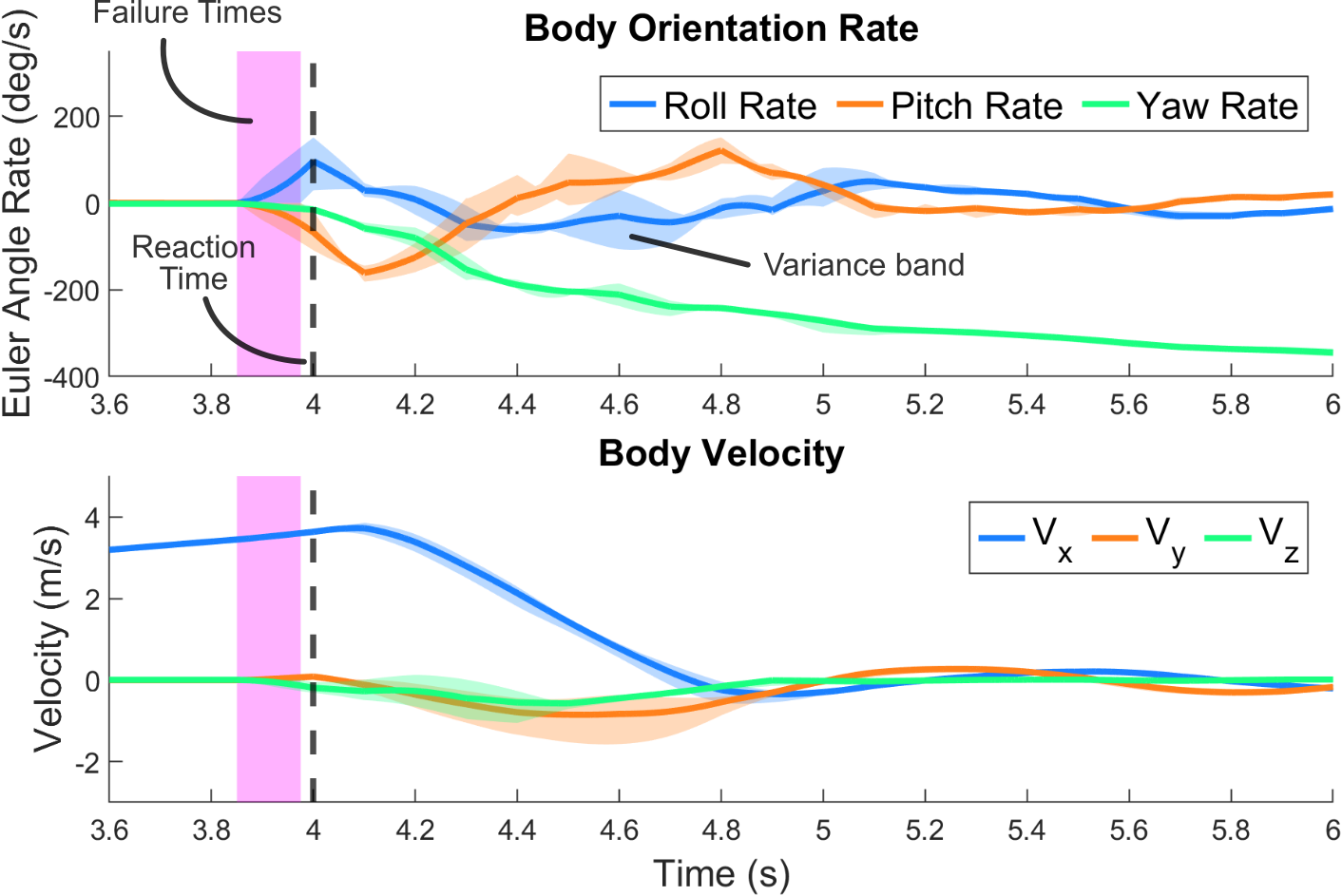}
    \caption{Plot showing the simulated body linear and angular velocities from the beginning with fully functional thrusters, thruster failure events, and recovery states. The yaw is unstable due to the loss of a thuster and the NMPC prioritizing roll and pitch stability during the failure recovery phase.}
    \label{fig::sa-sc-12}
\end{figure}

Figure~\ref{fig::sa-sc-11} and Figure~\ref{fig::sa-sc-12} present the robot's body position, Euler angles, body-frame velocities, and angular rates. Just prior to the failure, the robot travels at over 4\,m/s. Following the fault, it begins rotating about its yaw axis and settles into a new configuration within approximately 1\,s, with the yaw rate progressively increasing. This behavior is expected, as yaw is unconstrained in this configuration and the NMPC prioritizes recovery in roll and pitch.

\begin{figure}[!htbp]
    \centering
    \includegraphics[width=0.7\linewidth]{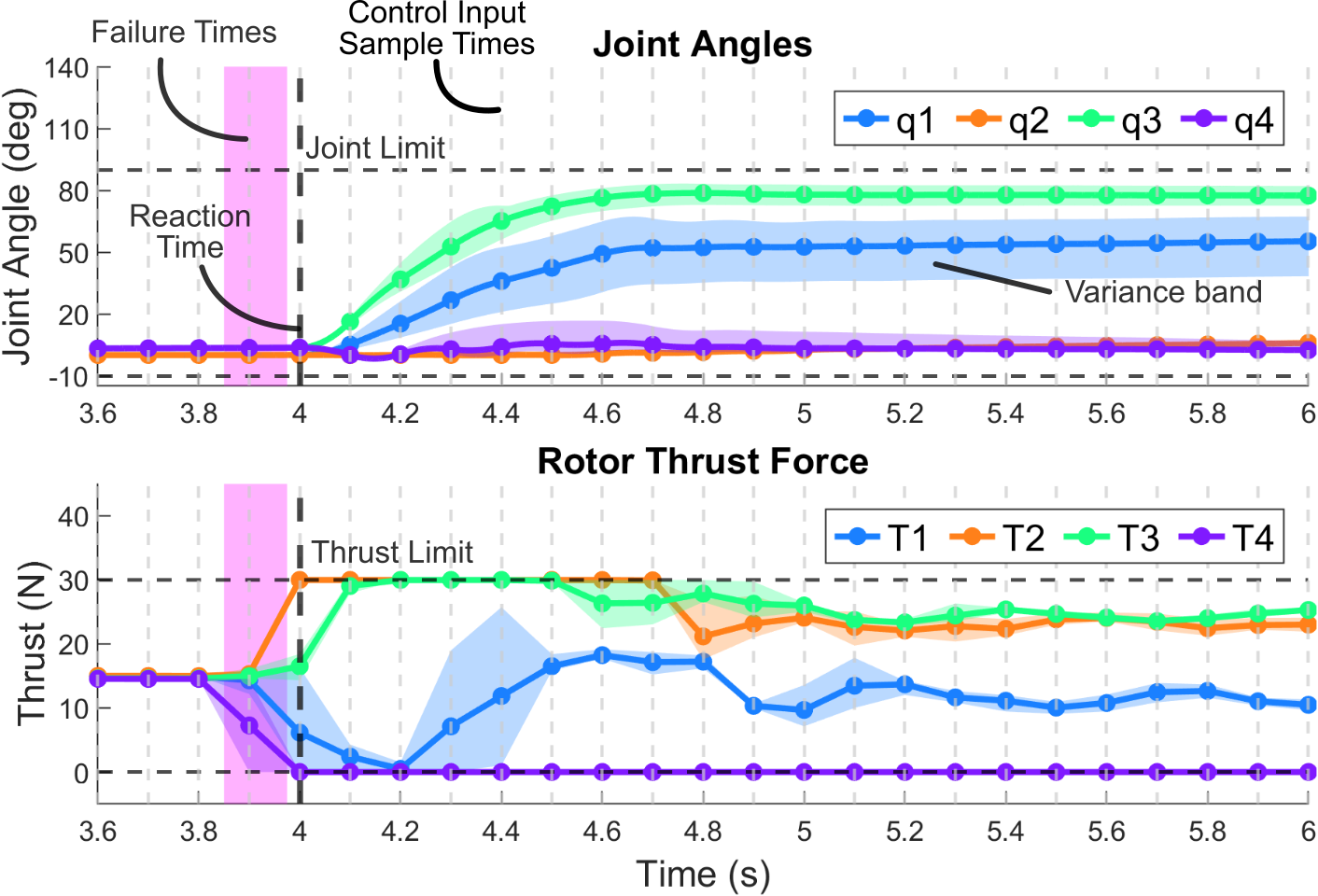}
    \caption{Plots showing the control inputs (joint angle and thruster values) in the simulation as calculated by the NMPC during the thruster failure events. Thruster 4 force goes to zero during the failure event.}
    \label{fig::sa-sc-13}
\end{figure}

Figure~\ref{fig::sa-sc-13} shows the control inputs, including thrust values and sagittal joint angles. The drop in thrust from rotor 4 confirms the failure event, while compensatory actions in the remaining actuators allow the robot to stabilize. Despite the lack of frontal joint actuation, the robot maintains control over its pitch and roll through coordinated use of sagittal joint movement and thrust vectoring.

\begin{figure}[!htbp]
    \centering
    \includegraphics[width=1\linewidth]{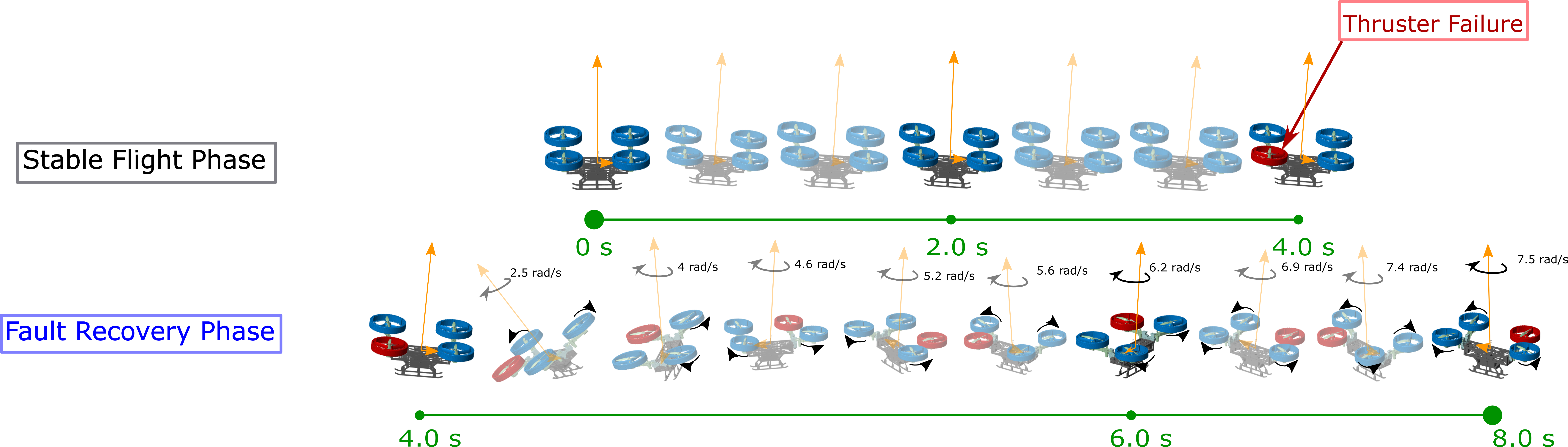}
    \caption{Illustrations of the robot during the flight, showing a stable flight without and with thruster failure (shown in red). The fault recovery phase and trajectory are shown in the bottom row.}
    \label{fig::sa-sc-14}
\end{figure}

To aid visualization, Figure~\ref{fig::sa-sc-14} provides sequential simulation snapshots depicting both the pre-failure stable flight phase and the post-failure recovery. The first row illustrates the robot’s forward flight posture up to 4\,s. After the failure, changes in body orientation are clearly visible. However, through the use of posture manipulation and coordinated thrust redistribution, the robot regains stability in roll and pitch, while accepting a steady yaw rotation as part of the new flight equilibrium.

These results collectively validate the NMPC framework’s capacity to handle abrupt actuator failure without requiring prior fault identification or reconfiguration, thereby underscoring its applicability in real-time scenarios.

\subsubsection{Scenario 2: Trajectory Tracking}

In this scenario, the robot is tasked with tracking a predefined trajectory consisting of multiple waypoints. The simulation is divided into several phases: a nominal stable flight phase, followed by progressively increasing Loss of Effectiveness (LoE) in rotor 4, culminating in complete failure and a final landing phase. Here, LoE is defined as the percentage reduction in thrust relative to the required hover thrust.

\begin{figure}[H]
    \centering
    \includegraphics[width=1\linewidth]{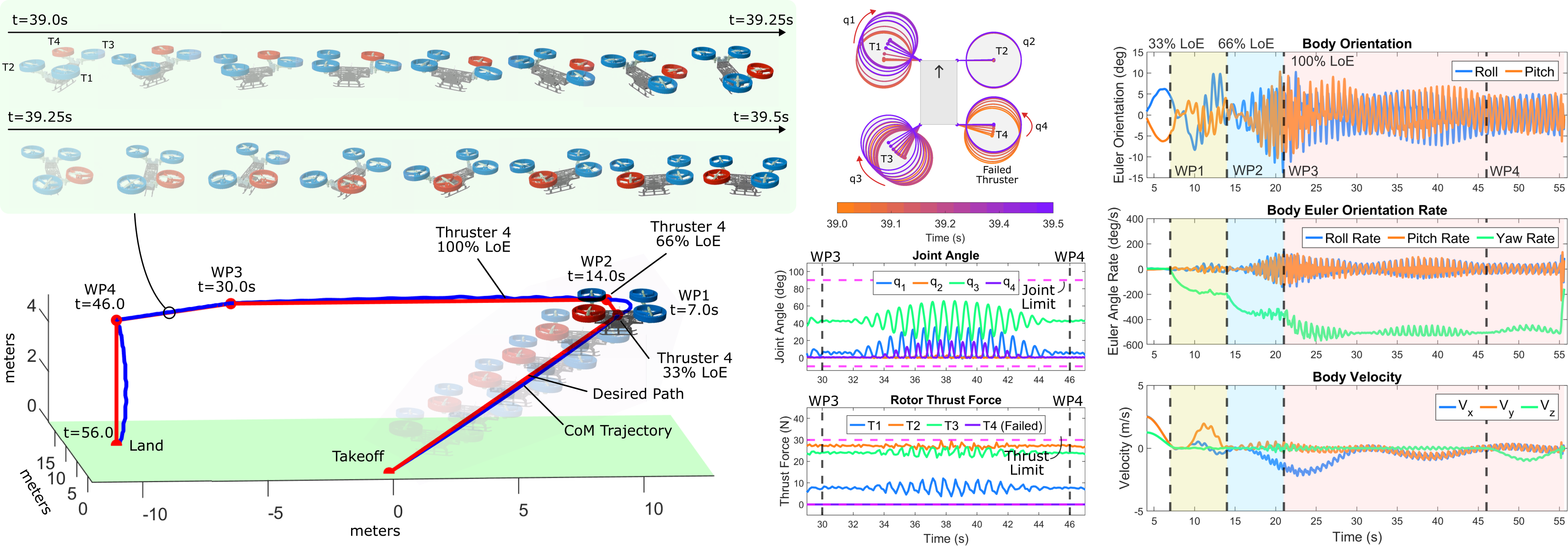}
    \caption{Plot showing the center of mass trajectory of the robot during the multiple failure events. The robot tracks the reference trajectory and hovers at each waypoint (WP1-WP4) despite the failure and successfully lands. Top middle shows posture manipulation in the body fixed frame used to achieve trajectory tracking despite thruster failure. This movement is performed periodically as the robot yaws until the next waypoint is reached, as shown in the graph below. Body orientation, orientation rate and velocity for the complete trajectory are also plotted.}
    \label{fig::sa-sc-21}
\end{figure}

The robot is allowed to stabilize for approximately 1~s at each waypoint before proceeding to the next segment of the trajectory. The performance under these failure conditions is illustrated in Fig.~\ref{fig::sa-sc-21}. As the LoE begins, the robot starts spinning about its primary axis, resulting in a gradually increasing yaw rate. After rotor 4 experiences a complete failure, the yaw rate reaches a saturated value within 10~s.

Despite the degraded thrust conditions, the robot maintains accurate trajectory tracking, as evident from the position plot in Fig.~\ref{fig::sa-sc-21}. This highlights the controller’s robustness in compensating for actuator faults while achieving the desired waypoints. During the final phase, the robot initiates a controlled descent along the $z$-axis, shutting down thrust and achieving a smooth landing as the yaw rate rapidly converges.

\begin{figure}[H]
    \centering
    \includegraphics[width=0.9\linewidth]{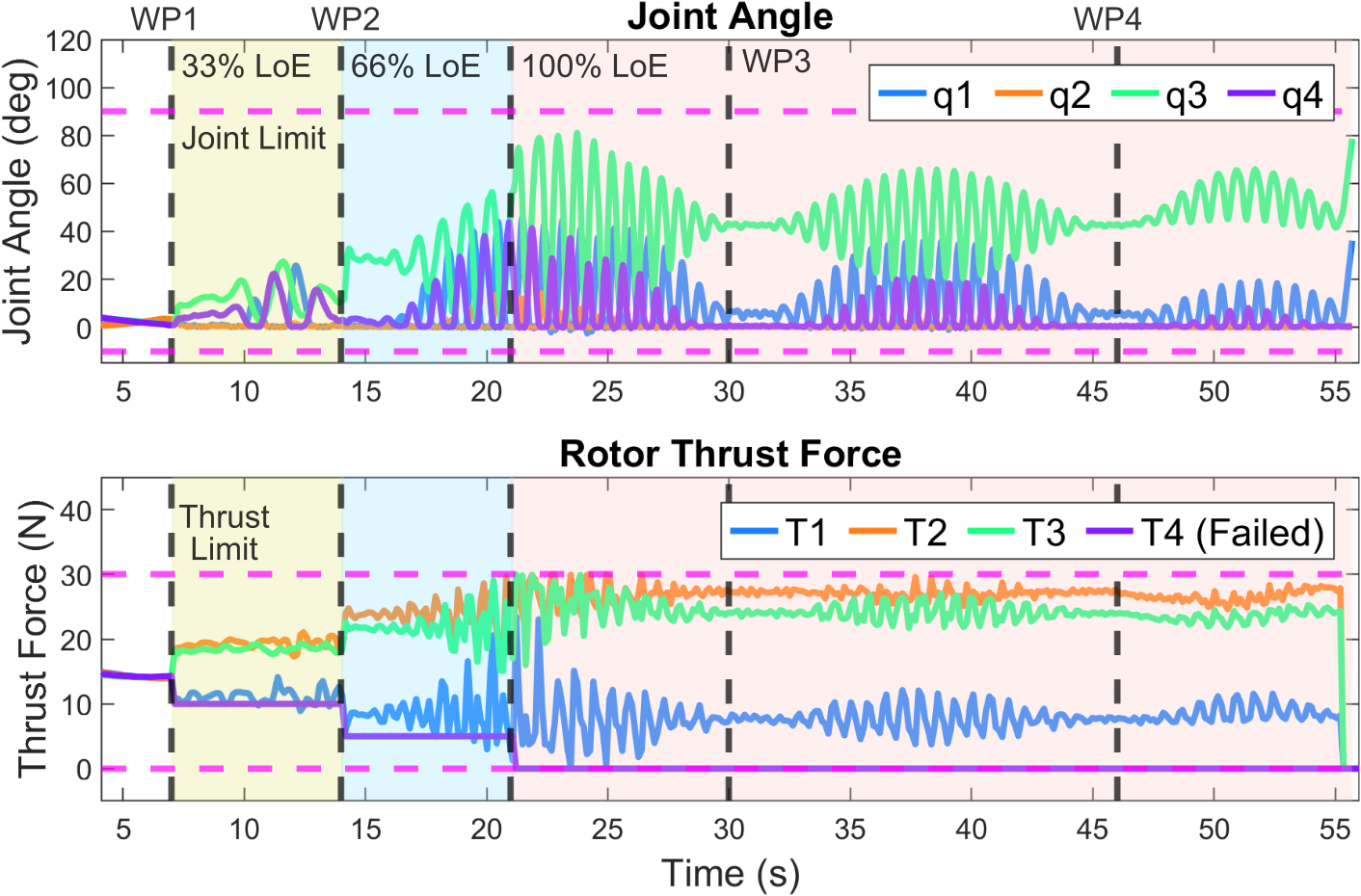}
    \caption{ Plot for thrust forces and joint angles for trajectory tracking during multiple loss of power events of varying severity. Rotor 4 undergoes 33\% LoE from 7 s to 14 s, 66\% LoE from 14 s to 21 s, and complete failure after 21 s. The plots show the evolution of the controller values through multiple degrees of thruster failures.}
    \label{fig::sa-sc-22}
\end{figure}

The corresponding thrust force commands and sagittal joint actuation profiles throughout this entire sequence are presented in Fig.~\ref{fig::sa-sc-22}. These plots highlight the controller’s adaptive behavior in response to varying severity of actuator failure.

\subsection{Simulation Results for Sagittal and Frontal-Actuated Joint Model}
In the sagittal and frontal actuated joint model, the controller is provided with actuation authority over both the sagittal and frontal hip joints. This allows the legs of the M4 robot to perform pronation and supination movements, towards and away from the body, in addition to sagittal motion. Compared to the sagittal-only configuration, this expanded actuation capability enhances maneuverability and control authority during fault recovery.

In this framework, the yaw angle is not explicitly constrained; however, deviations in yaw are penalized with a higher cost in the NMPC objective function relative to the sagittal-only configuration. This encourages yaw stability while preserving the robot's ability to adapt to fault-induced dynamics.

Simulation results for this model are evaluated in two scenarios: (1) fault recovery during hover, and (2) fault recovery during trajectory tracking. The following sections present controller performance in both cases, highlighting the improved ability of the system to recover from actuator failures, particularly in stabilizing the robot with respect to yaw dynamics.

\subsubsection{Scenario-1: Fault Recovery During Hover} 

In this scenario, the robot is commanded to maintain a stationary hover position while subjected to a sequence of actuator failures. The objective is to evaluate the system’s ability to preserve flight stability and orientation through partial and complete loss of thrust from a single rotor.

\begin{figure}[!htbp]
    \centering
    \includegraphics[width=1\linewidth]{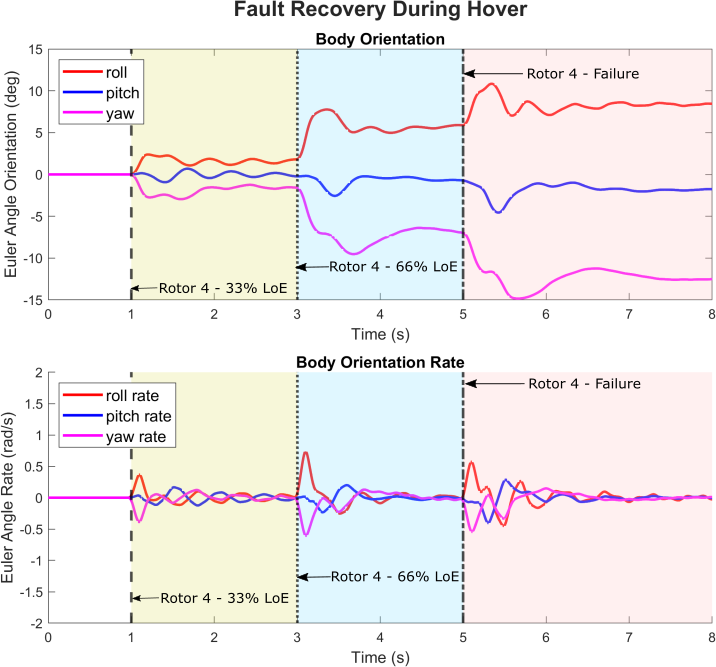}
    \caption{Plots showing the body orientation and orientation rate of the robot as it subsequently undergoes partial and complete thruster failure of rotor 4 during hover. The simulation result presented for a fully actuated model (2 DoF per arm) shows the capability to eliminate yaw rate even after complete thruster failure. Rotor 4 undergoes 33\% LoE from 1 s to 3 s, 66\% LoE from 3 s to 5 s, and complete failure after 5 s.}
    \label{fig::sfa-sc-11}
\end{figure}

A 33\% Loss of Effectiveness is introduced in rotor 4 at $t = 1$~s, followed by a 66\% LoE at $t = 3$~s, and a complete failure after $t = 5$~s. The resulting Euler angles and angular velocity responses are shown in Fig.~\ref{fig::sfa-sc-11}. It can be observed that each failure event leads to a transient deviation in body orientation, particularly in roll and yaw. However, the system is able to re-stabilize after each disturbance. Notably, unlike the sagittal-only actuated model, the robot does not exhibit an increasing yaw rate after complete failure, indicating improved yaw stability due to the additional frontal joint actuation.

\begin{figure}[!htbp]
    \centering
    \includegraphics[width=1\linewidth]{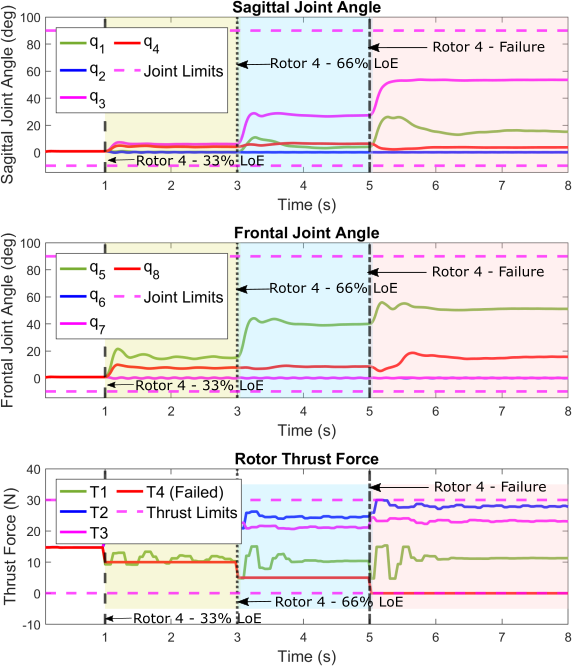}
    \caption{Plots showing the control inputs (joint angles and thruster values) in the simulation as calculated by the NMPC during the partial and complete thruster failure of rotor 4 for a fully actuated model. Rotor 4 undergoes 33\% LoE from 1 s to 3 s, 66\% LoE from 3 s to 5 s, and complete failure after 5 s.}
    \label{fig::sfa-sc-12}
\end{figure}

The corresponding control inputs, thrust forces and joint angles, are presented in Fig.~\ref{fig::sfa-sc-12}, illustrating the controller's adaptive response to the failure conditions. These results demonstrate that the fully actuated joint configuration enables the robot to rapidly stabilize its orientation and maintain controlled flight, even after the complete loss of a rotor.

\subsubsection{Scenario-2: Trajectory Tracking}

In this scenario, the robot is tasked with tracking a predefined trajectory composed of multiple waypoints to evaluate its performance under actuator faults during transient flight conditions. The simulation is divided into several phases: an initial nominal stable flight, followed by progressively increasing LoE in rotor 4, and concluding with a final descent phase to reach the goal position.

\begin{figure}[!htbp]
    \centering
    \includegraphics[width=1\linewidth]{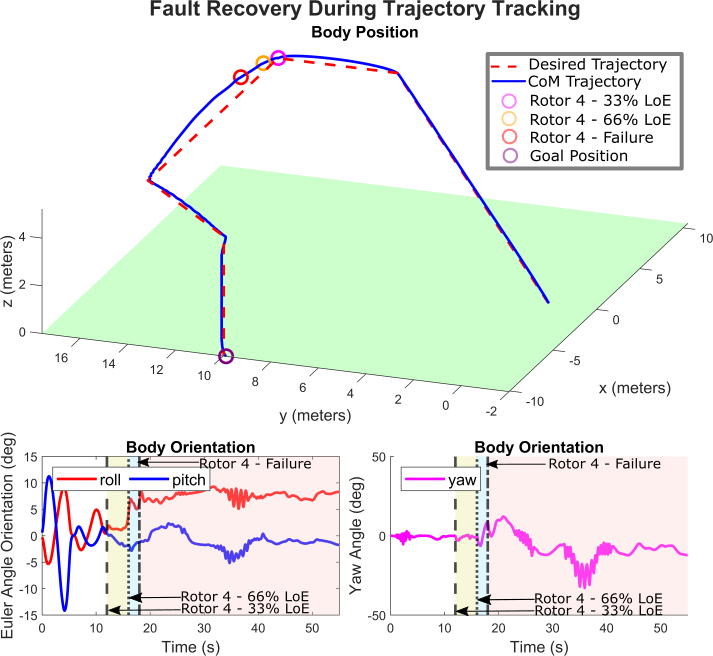}
    \caption{Plot showing the center of mass trajectory of the robot during the multiple failure events. The robot tracks the reference trajectory and hovers at each waypoint despite the failure and successfully reaches the goal position without constant spinning. Body orientation for the complete trajectory is also presented.  Rotor 4 undergoes 33\% LoE from 12 s to 16 s, 66\% LoE from 16 s to 18 s, and complete failure after 18 s.}
    \label{fig::saf-sc-21}
\end{figure}

The robot is allowed to stabilize for approximately 1~s at each waypoint before proceeding to the next segment of the trajectory. In the initial ascent phase, the robot rapidly moves from the first waypoint to the second. A 33\% LoE is introduced at $t = 12$~s, followed by a 66\% LoE at $t = 16$~s, and complete failure of rotor 4 after $t = 18$~s. The robot’s performance under these failure conditions is shown in Fig.~\ref{fig::saf-sc-21}. As the LoE progresses, the controller dynamically adjusts the thrust values as well as the sagittal and frontal joint angles to compensate for the actuator degradation and maintain stability while tracking the desired trajectory.

Despite the compromised thrust availability, the robot successfully maintains accurate tracking of the reference trajectory, as evident from the position plots in Fig.~\ref{fig::saf-sc-21}. This demonstrates the robustness of the NMPC controller in handling severe actuator faults while achieving waypoint objectives. During the final phase, the robot performs a controlled descent along the $z$-axis, smoothly reaching the target goal position.

\begin{figure}[!htbp]
    \centering
    \includegraphics[width=1\linewidth]{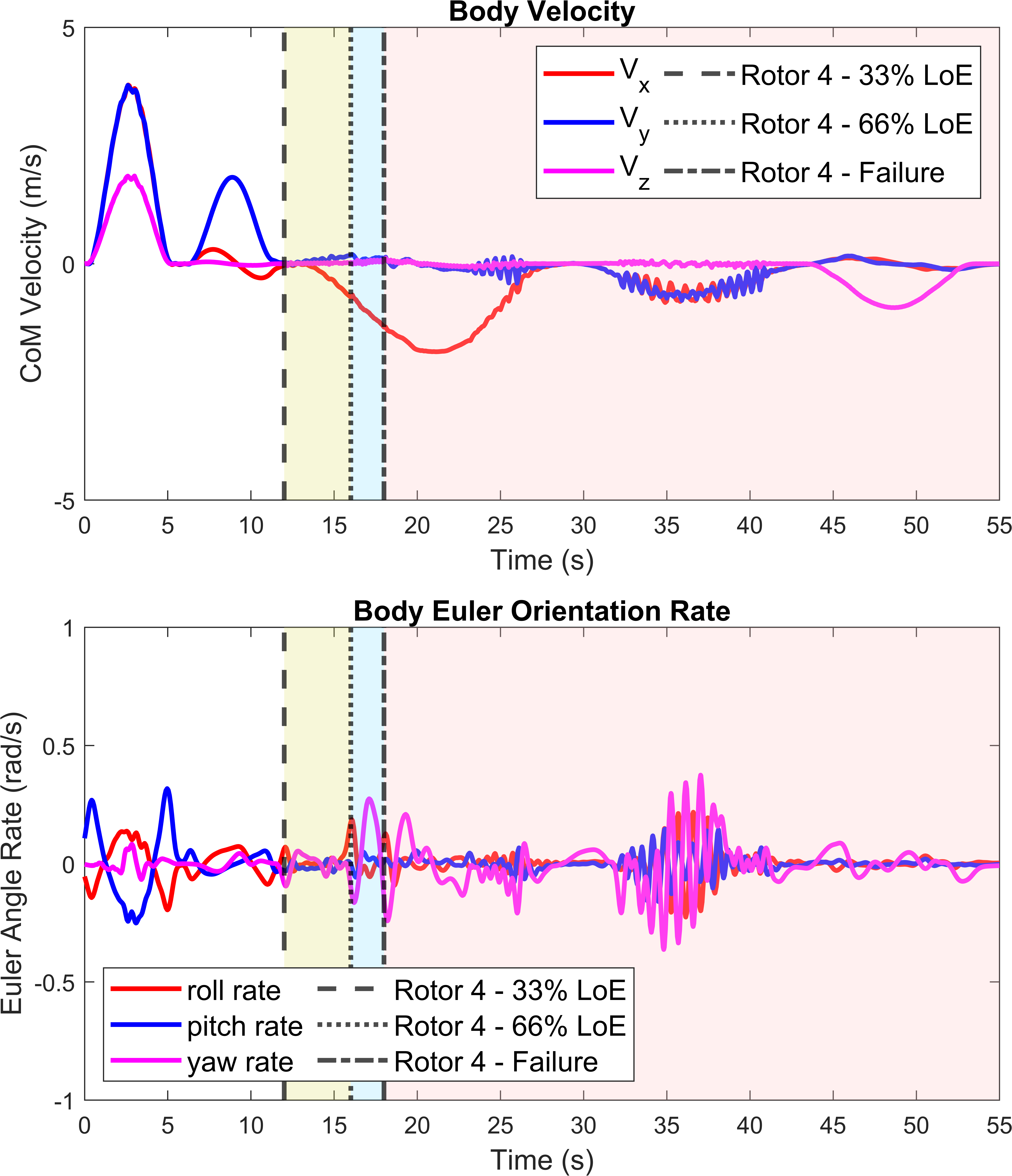}
    \caption{Plots showing the center of mass velocity and orientation rate throughout the course of trajectory tracking. Rotor 4 undergoes 33\% LoE from 12 s to 16 s, 66\% LoE from 16 s to 18 s, and complete failure after 18 s.}
    \label{fig::saf-sc-22}
\end{figure}

\begin{figure}[!htbp]
    \centering
    \includegraphics[width=1\linewidth]
    {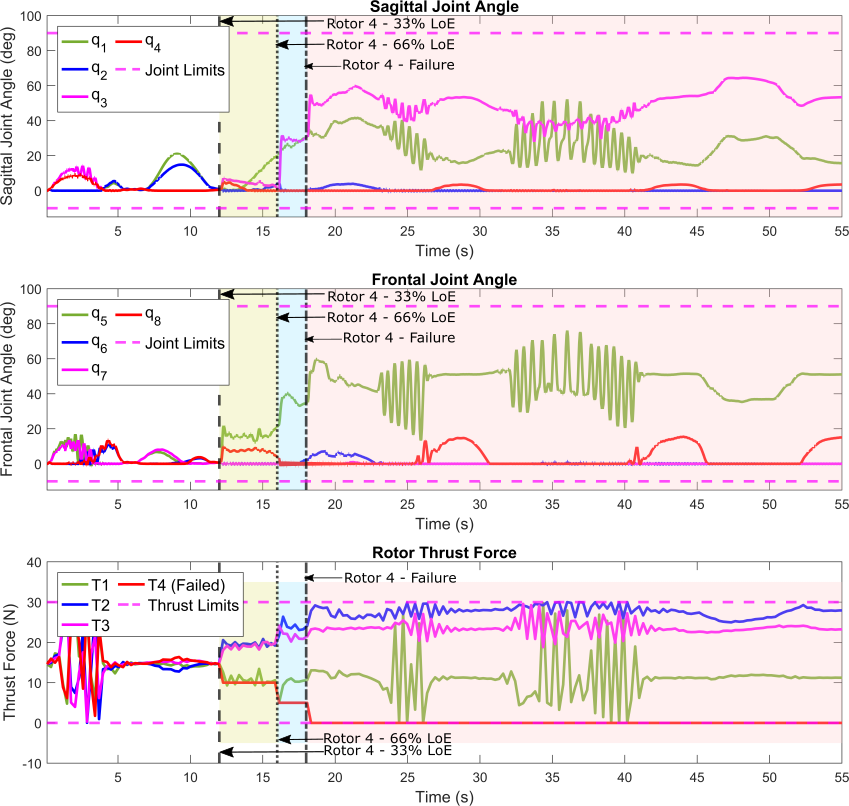}
    \caption{Plots showing the control inputs (joint angles and thruster values) in the simulation as calculated by the NMPC during the partial and complete thruster failure of rotor 4 during trajectory tracking. It can seen at the instances of LoE's and failure that joint angles and thrust forces are rapidly adjusted to accommodate thruster failure of rotor 4.Rotor 4 undergoes 33\% LoE from 12 s to 16 s, 66\% LoE from 16 s to 18 s, and complete failure after 18 s.}
    \label{fig::saf-sc-23}
\end{figure}

The corresponding thrust force commands, along with sagittal and frontal joint angle trajectories throughout the entire sequence, are shown in Fig.~\ref{fig::saf-sc-22} and Fig.~\ref{fig::saf-sc-23}. These results highlight the controller’s adaptive response to varying severity of thruster failure and its ability to maintain stable flight and tracking performance without inducing constant spinning or instability.

\section{Agile Trajectory Tracking}
\label{sec:agile_tracking}

The proposed NMPC controller was further evaluated for its ability to enable highly agile flight maneuvers at elevated speeds. In this experiment, the M4 robot was tasked with executing sharp turns of varying angles, specifically $30^\circ$, $60^\circ$, $90^\circ$, and $120^\circ$, while maintaining forward speeds exceeding $10\,\mathrm{m/s}$. This set of maneuvers was designed to simulate dynamic flight through cluttered environments, such as dense forests or urban canyons, where rapid reorientation is critical for obstacle avoidance and trajectory adaptation.

\begin{figure}[H]
    \centering
    \includegraphics[width=1\linewidth]
    {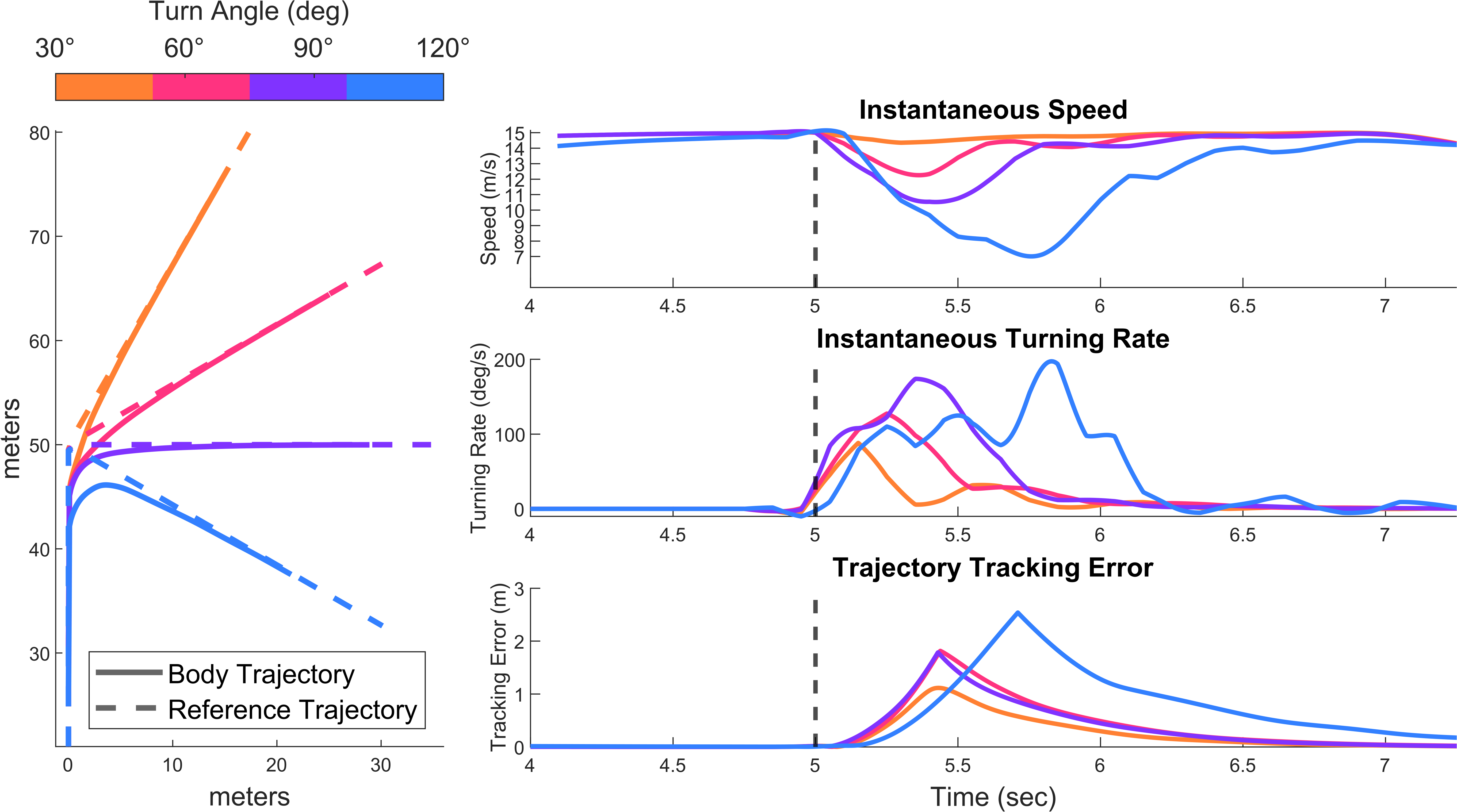}
    \caption{Position tracking, speed, turning rate, and tracking error during agile flight maneuvers with turning angles of $30^\circ$, $60^\circ$,$90^\circ$, 
    and $120^\circ$.}
    \label{fig::att-sc-11}
\end{figure}

Figure~\ref{fig::att-sc-11} presents a comprehensive view of the robot’s performance during these maneuvers. The left panel of the figure shows the spatial trajectories for each turn, overlaid with their respective reference paths. Across all cases, the M4 robot consistently tracks the intended trajectories with minimal lateral deviation. For smaller turn angles, such as $30^\circ$ and $60^\circ$, the robot exhibits tight adherence to the path, with only marginal offsets during the turn execution. As the turn angle increases to $90^\circ$ and $120^\circ$, the demands on lateral acceleration grow substantially, leading to slightly larger deviations from the reference, but without any loss of overall stability or control.

The speed profiles for each maneuver, shown in the top-right plot, reveal the dynamic behavior of the system during high-speed flight. Prior to each turn, the robot maintains a cruising speed around $14.5\,\mathrm{m/s}$. As it approaches the turning waypoint, a controlled deceleration is observed, proportional to the sharpness of the turn. In the most aggressive $120^\circ$ case, the speed temporarily drops below $8\,\mathrm{m/s}$, reflecting the greater thrust redistribution and posture adjustments needed to maintain stability during the maneuver. Following the turn, the system smoothly reaccelerates to regain forward momentum.

The middle-right plot captures the turning rate, highlighting the rapid reorientation achieved by the robot. Sharp peaks corresponding to the turn initiations are clearly visible, with peak yaw rates increasing with the turn severity. For the $120^\circ$ turn, the yaw rate exceeds $200^\circ/\mathrm{s}$, indicating a highly aggressive maneuver that is nonetheless executed without overshoot or instability, thanks to the controller’s coordinated manipulation of joint articulation and thrust.

Finally, the bottom-right plot depicts the trajectory tracking error over time. Even during aggressive maneuvers, the tracking error remains tightly bounded. For the smaller turns, the error remains under $1\,\mathrm{m}$ throughout the maneuver. In the more challenging $90^\circ$ and $120^\circ$ cases, transient error peaks of approximately $2$ to $2.5\,\mathrm{m}$ are observed during the turning phase, but these are rapidly corrected as the robot exits the turn. 

Overall, the results demonstrate the NMPC framework’s effectiveness in enabling agile, high-speed trajectory tracking using the combined authority of thrust vectoring and posture manipulation. The M4 robot is able to dynamically negotiate sharp turns with minimal trajectory deviation, maintaining stability and control throughout aggressive maneuvers, without any requirement for preplanned maneuver-specific tuning or controller switching.

 \chapter{Conclusion}
\label{chap:conclusion}

This thesis presents a unified control framework for agile and fault-tolerant flight of the Multi-Modal Mobility Morphobot through the integration of posture manipulation and thrust vectoring. The proposed approach leverages the redundancy provided by M4's articulated legs to compensate for partial or complete actuator failures in aerial mode. A nonlinear model predictive control architecture is developed using a reduced-order dynamic model as a prediction model, enabling the system to perform online trajectory tracking and fault recovery with real-time feasibility.

To validate the controller, a high-fidelity simulation environment was created using Simscape Multibody, where the robot's complete dynamics, including aerodynamic effects and actuator configurations, were modeled. Multiple simulation scenarios were tested to evaluate the system's fault recovery capabilities: first, with only hip-sagittal actuation and second, with both sagittal and frontal hip joints actuated. In both cases, the controller successfully stabilized the robot during forward flight, hovering, and trajectory tracking tasks under conditions of progressively worsening rotor effectiveness.

The results clearly demonstrate that the proposed NMPC controller is capable of robust recovery without requiring explicit fault detection or controller switching. In particular, the fully actuated configuration (sagittal + frontal) exhibited superior performance in rejecting yaw disturbances and maintaining stability under severe failure conditions. Furthermore, the same NMPC framework enabled high-speed, agile trajectory tracking, demonstrating the system's versatility and adaptability. The controller was able to execute rapid maneuvers along sharp trajectories, including turns of up to $120^\circ$, without compromising stability.

In summary, this work contributes a robust and flexible control strategy that unifies posture manipulation and thrust vectoring to enhance both agility and fault tolerance in morphing aerial robots. The methodology and insights from this research pave the way for the deployment of resilient multi-modal platforms in challenging and uncertain environments such as planetary exploration or disaster response.

\section{Future Work}

This thesis lays the groundwork for a unified NMPC-based control strategy combining posture manipulation and thrust vectoring. However, several promising research directions remain:

\begin{itemize}
    \item \textbf{Hardware Implementation and Real-Time Validation:} The current framework has only been validated in simulation. A critical next step is the deployment of the NMPC controller on the physical M4 platform. This would require real-time optimization on edge devices and robust state estimation under sensor noise and model mismatch.

    \item \textbf{Experimental Fault Tolerance:} Validating the controller's robustness against real actuator degradations, such as latency, saturation, and partial failures, through hardware experiments is essential to demonstrate its practical fault-tolerance capability.

    \item \textbf{Fault Detection and Isolation (FDI):} Although the current framework operates without explicit fault detection, integrating a lightweight FDI module could enable early-stage identification of faults and further enhance recovery strategies through adaptive reconfiguration.

    \item \textbf{Perception-Aware and Dynamics-Aware Planning:} Future work could include integrating onboard perception (e.g., vision or LiDAR) into the NMPC loop to enable obstacle-aware and terrain-aware maneuvering. Additionally, embedding dynamics-aware planning could facilitate safe navigation through cluttered environments like forests or collapsed infrastructure.

    \item \textbf{Learning-Based Surrogates and Model Adaptation:} Incorporating learning-based surrogate models or data-driven adaptation mechanisms could improve prediction accuracy and reduce computation time. This would support adaptation to actuator wear, component aging, or environmental disturbances.

    \item \textbf{Multi-Modal Integration:} Extending the NMPC framework to enable control during transitions between aerial, wheeled, and legged locomotion would achieve true multi-modal autonomy. This would support real-world missions where robots must adapt locomotion modes dynamically.
\end{itemize}

These directions collectively aim to enhance the autonomy, adaptability, and field readiness of morphing aerial platforms like the M4 for planetary exploration, disaster response, and search-and-rescue applications.

\bibliographystyle{IEEEtran}  

\bibliography{bib/thesis}

\appendix


\end{document}